\documentclass[journal]{IEEEtran}
\ifCLASSINFOpdf
\else
\fi

\usepackage{amsmath}
\usepackage[table]{xcolor}
\usepackage{graphicx,multirow,tikz}
\usepackage{amsfonts}
\usepackage[colorlinks,citecolor=green,urlcolor=magenta,bookmarks=false,hypertexnames=false]{hyperref}
\begin{document}

\title{R-PointHop: A Green, Accurate, and Unsupervised Point Cloud
Registration Method}

\author{Pranav~Kadam,~\IEEEmembership{Student Member,~IEEE,}
        Min~Zhang,
        Shan~Liu,~\IEEEmembership{Senior Member,~IEEE,}
        and~C.-C.~Jay~Kuo,~\IEEEmembership{Fellow,~IEEE}
\thanks{Pranav Kadam is with the Ming Hsieh Department of Electrical and Computer
Engineering, University of Southern
California, CA, 90089, USA, e-mail: prIEEEanavka@usc.edu. }
\thanks{Min Zhang is with the Ming Hsieh Department of Electrical and Computer
Engineering, University of Southern
California, CA, 90089, USA, e-mail: zhan980@usc.edu. }
\thanks{Shan Liu is with Tencent Media Lab, Tencent America, 2747 Park
Blvd, Palo Alto, CA, 94306 USA, email: shanl@tencent.com. }
\thanks{C.-C. Jay Kuo is with the Ming Hsieh Department of Electrical and Computer
Engineering, University of Southern
California, CA, 90089, USA, e-mail: cckuo@sipi.usc.edu. }}

\maketitle

\begin{abstract}

Inspired by the recent PointHop classification method, an unsupervised
3D point cloud registration method, called R-PointHop, is proposed in this work.
R-PointHop first determines a local reference frame (LRF) for every
point using its nearest neighbors and finds local attributes.  Next,
R-PointHop obtains local-to-global hierarchical features by point
downsampling, neighborhood expansion, attribute construction and
dimensionality reduction steps. Thus, {point} correspondences {are built} in hierarchical feature space using the nearest neighbor
rule. Afterwards, a subset of salient points {with} good correspondence is
selected to estimate the 3D transformation. The use
of {the} LRF allows for {invariance of the} hierarchical features of points with
respect to rotation and translation, thus making R-PointHop more robust {at} building point correspondence, even when the rotation angles are large.
Experiments are conducted on the 3DMatch, ModelNet40, and Stanford Bunny
datasets, which demonstrate the effectiveness of R-PointHop for 3D
point cloud registration. R-PointHop’s model size and training time are an order of magnitude smaller than those of deep learning methods, and its registration errors are smaller, {making it a green and accurate solution}. Our codes are available on GitHub
\footnote{\url{https://github.com/pranavkdm/R-PointHop}}. 

\end{abstract}


\IEEEpeerreviewmaketitle

\section{Introduction}\label{sec:intro}

\IEEEPARstart{R}{egistration} is a key step in many applications
{of} point clouds. Given a pair of point cloud scans, registration
attempts to find a rigid transformation {for their optimal alignment}. Multiple point cloud scans {can be} registered to get a complete 3D
scene of the environment. With the rapid development reduction in cost of 3D scanning
devices such as LiDAR, point cloud
processing has been on the rise. The quality of registration directly
affects downstream tasks, including classification, segmentation, object
detection, pose estimation, and odometry. These tasks are commonly
encountered in autonomous driving, robotics, computer graphics,
localization, AR/VR, and so on. Point cloud registration {is} an active
research topic. {The current} focus is on development of
learning models for registration that can handle challenges such as noise, varying point densities, outliers, occlusions, and
partial views. 

The correspondence problem exists in quite a few computer vision tasks.
From the viewpoint of 2D images, the interest lies in finding matching
pixels or regions {between} multiple images. These correspondences can be
used for image stitching to generate a panorama
\cite{brown2007automatic, juan2010surf}, 3D reconstruction, or structure
from motion (SfM) \cite{geiger2011stereoscan, mouragnon2006real,
schonberger2016structure}. 2D descriptors are often extracted {using} the
scale-invariant feature transform (SIFT) \cite{lowe2004distinctive} or speeded-up robust features (SURF) \cite{bay2006surf} {algorithms}. These methods are
effective in building correspondence {between} pixels of different images.
Similarly, in the context of 3D point clouds, geometric registration
based on point correspondence {is} popular. {The most common methods include} the classical
iterative closest point (ICP) \cite{besl1992method} and its
derivatives \cite{chen1992object}, \cite{segal2009generalized}.
Correspondence-aided odometry and mapping {has also been} demonstrated for 3D point
clouds \cite{zhang2014loam}. 

To develop an accurate 3D correspondence solution, it is essential to {achieve} good {point} feature representation. Desirable properties for
point features include: 1) robustness to noise, outliers and the point
density, 2) invariance under rigid motion, and 3) global context
awareness. Earlier solutions, e.g., \cite{tombari2010unique},
\cite{johnson1997spin}, {have} used 3D descriptors to capture local geometric
properties such as surface normals, tangents and curvatures.  These
primitive descriptors are derived based on the first- or higher-order
statistics of neighboring points, histogram, angles, etc.  The recent
trend is to learn features from an end-to-end optimization setting with
deep neural networks (DNNs) and build correspondences accordingly. To
this end, we propose a new method, called R-PointHop\footnote{The
acronym indicates a point cloud registration method built upon features
learned by PointHop \cite{zhang2020pointhop} or
PointHop++\cite{zhang2020pointhop++}}, to learn features in an
unsupervised manner for point correspondence. These
correspondences are {then} used to find the 3D transformation for
registration. 

Supervised learning via DNNs has revolutionized the field of 3D point
cloud processing. PointNet \cite{qi2017pointnet} {is} the first well
known DNN solution that uses learned features for point cloud
classification and semantic segmentation. Several follow-up {works}
\cite{qi2017pointnet++}, \cite{wang2019dynamic}, \cite{li2018pointcnn}
{have reinforced} the belief that large-scale point cloud processing can benefit
from deep learning.  Researchers have developed a large number of DNNs
for various 3D vision tasks {including} correspondence
\cite{zeng20173dmatch}, \cite{deng2018ppfnet}, \cite{deng2018ppf} and
registration \cite{wang2019deep, choy2020deep, aoki2019pointnetlk}.
These methods formulate registration as a supervised learning problem
and solve it using end-to-end optimization. 

Different {forms of} supervision have been adopted in deep learning,
including ground truth transformations, valid and invalid
correspondence {pairs}, and object labels. On one hand, it is difficult and/or
expensive to {obtain} these labels in real world applications, {while} on the other, unsupervised learning and model-free methods fail to match the 
performance of deep learning methods, especially for complex point cloud sets. 
A question of interest is whether the performance gain is due to a 
large number of unlabeled data, data labeling, or both of them. 
Other concerns in real world applications are model complexity 
(in terms of memory requirement) and computational complexity (in terms
of training/inference time). Deep learning methods often run on 
GPUs since they demand larger model sizes and longer training/inference 
time. Point cloud processing using deep learning is no exception. 
It is {desirable} to look for a green solution that {consumes} much less power. This implies a {method with a} smaller model size and less training/inference 
time, yet, {whose} performance {is} on par with that of DNNs. 

{Two} green point cloud classification methods {have been} proposed before, {namely} the PointHop method \cite{zhang2020pointhop} and the
PointHop++ method \cite{zhang2020pointhop++}. Both {methods} extract point cloud
features in a one-pass feedforward manner without any label information.
These features were fed into a classifier {such as} random forest (RF) or support vector machine (SVM) for point cloud classification.  The {salient points analysis} SPA
method \cite{kadam2020unsupervised} {extends} PointHop++ for 3D
registration. These methods are designed based on the successive
subspace learning (SSL) framework. SSL offers a promising direction for
point cloud research {due to} its interpretability, small model size, low
training/inference time, and good performance. PointHop
and PointHop++ assume that objects are aligned in a canonical {frame}
before processing. This assumption does not hold in general and 3D
registration is usually needed as a pre-processing step. Note that SPA
may fail to align two point clouds that are related to each other with
larger rotation angles, since it derives its features using PointHop++. 

To address the shortcomings of SPA, R-PointHop offers a new way {of extracting point} features that are invariant to point cloud rotation and
translation. Rotation invariance is achieved by considering a local
reference frame (LRF) defined at each point. This enables R-PointHop to
find robust point correspondences even when the rotation angle is large.
Besides, R-PointHop covers partial-to-partial registration which is
often encountered in real world problems. In contrast, SPA does not
account for partial registration, thus, limiting its application scope.
Furthermore, it is observed that point cloud features of similar local
structures are clustered closely in the feature space. This indicates
that R-PointHop features could be used as 3D local descriptors and
applied to a wide range of tasks that go beyond 3D registration. 

The main contributions of this work are summarized below.
\begin{itemize}
\item An unsupervised feature learning method, called
R-PointHop is proposed. R-PointHop learns point features that are invariant {to} point cloud rotation and translation.
\item The effectiveness of {the} proposed features for geometric registration task is demonstrated through {a} series of
experiments on indoor point cloud scans as well as synthetic and
real-world models.
\item Emphasis is given {to} {designing} a green
solution that {has a smaller model size, lower memory consumption, and reduced training time} as compared to state-of-the-art methods.
\end{itemize}

The rest of this {paper} is organized as follows. Related work is reviewed
in Sec. \ref{sec:review}, where both model-free and
learning-based methods for 3D correspondence are examined. The SSL
framework, which forms the basis for R-PointHop, is also discussed. The
R-PointHop method is proposed in Sec. \ref{sec:method}. It consists of
the local reference frame (LRF) computation, attribute construction, and
multi-hop feature learning. Correspondence selection is also discussed.
Experimental results on the 3DMatch \cite{zeng20173dmatch}, ModelNet40 \cite{wu20153d} and Stanford Bunny
dataset \cite{turk1994zippered, curless1996volumetric,
krishnamurthy1996fitting} are reported in Sec. \ref{sec:experiments}. In Sec. \ref{sec:discussion}, additional discussion
on the role of supervision in the point cloud registration is provided.
We also {examine} the limitation of
R-PointHop. Finally, concluding remarks are given in Sec.
\ref{sec:conclusion}. 

\section{Review of Related work}\label{sec:review}

\subsection{Classical Model-Free Methods}\label{subsec:model-free}

Classical registration methods such as the iterative closest point (ICP)
method and its variants (e.g., Point-to-plane ICP \cite{chen1992object},
Generalized-ICP \cite{segal2009generalized}, Go-ICP \cite{yang2015go},
etc.) have been used in point cloud registration for a long while. For
every point in one point cloud, ICP first finds its closest point in the
other point cloud. Then, point correspondences are used to estimate the
transformation that minimizes the mean squared error between the 3D
coordinates of matched points. Since ICP is local by nature, it works
well only when the optimal transformation is close to the initial
alignment. Go-ICP uses a Branch-n-Bound (BnB) module to search for a
globally optimal solution. Various modifications of ICP are summarized
and compared in \cite{rusinkiewicz2001efficient}. The Fast Global
Registration method \cite{zhou2016fast} conducts global registration of
partially overlapping surfaces without an initial alignment. 
It uses FPFH \cite{rusu2009fast} feature. {Meanwhile,} Teaser 
\cite{teaser, yang2019polynomial} uses truncated least squares to handle
scale, rotation and translation. The above-mentioned methods are
model-free. They use handcrafted features and solve an optimization
problem. 

\subsection{Local Geometric Descriptors}\label{subsec:local}

SIFT \cite{lowe2004distinctive} and SURF \cite{bay2006surf} are well
known 2D keypoint descriptors. Similarly, some local geometric
properties (e.g., eigen decomposition, surface normals, signatures,
curvatures, histograms, and angles) can be used to describe points in 3D
point clouds. SHOT \cite{tombari2010unique} is a 3D descriptor based on
the histogram of point normals in a local support region. Spin-images
\cite{johnson1997spin} is a local surface representation comprising of
oriented points and their images. FPFH \cite{rusu2009fast} combines 3D
coordinates and surface normals of $k$ nearest neighbors of a point.
USC \cite{tombari2010usc} is a modification of SHOT.  The initial idea
of R-PointHop was inspired by these unsupervised local
descriptors. However, two additional ideas are added to make the local
geometrical descriptors more powerful. First, the target descriptor {is} learned from training samples rather than being defined by a
set of pre-determined rules. Second, it {has} multi-scale
representation capability. To meet the first criterion, we introduce {principal} component analysis (PCA) for feature extraction, which is data
driven. To meet the second criterion, features in R-PointHop are learned
in a multi-hop manner, where the neighborhood size grows as the
number {of hops} increases. This allows the derivation of multi-scale descriptors
centered at a point so that the short-, mid- and long-range neighborhood
information can be captured simultaneously. 

\subsection{Deep Learning Methods}\label{subsec:supervised}

{Several} deep learning methods have been proposed for point cloud classification,
segmentation and registration tasks in recent years.  PointNet
\cite{qi2017pointnet}, PointNet++ \cite{qi2017pointnet++} and DGCNN
\cite{wang2019dynamic} are well known networks in this field, and most
learning-based registration methods use them as the backbone.  Deep
Closest Point (DCP) \cite{wang2019deep} exploits point features learned
from DGCNN and uses a transformer to learn contextual information
between features of two point clouds to be registered. A differentiable
SVD module is designed to predict the rotation in an end-to-end manner.
PR-Net \cite{wang2019prnet} extends DCP for registration of partial
point clouds through an action-critic closest point module.  PointNetLK
\cite{aoki2019pointnetlk} uses the globally pooled features learned by
PointNet and employs the Lucas-Kanade (LK) algorithm
\cite{lucas1981iterative} to conduct registration in an iterative
manner. In contrast with DCP, PointNetLK does not demand explicit point
correspondences and {instead uses} an iterative LK algorithm. Both DCP
and PointNetLK use ground truth rotation {matrices} and translation
vectors to train end-to-end {networks}. Another approach is to
optimize an end-to-end network {that} finds 3D correspondences, where
supervision is provided in terms of valid and invalid correspondence
pairs. CORSAIR \cite{zhao2021corsair} combines
global shape embedding with local point-wise features to simultaneously
retrieve and register point cloud objects. 3DMatch
\cite{zeng20173dmatch} uses point correspondences available from RGBD
reconstruction datasets to train a siamese 3D CNN.  PPFNet
\cite{deng2018ppfnet} finds a local point pair feature embedding, which
is followed by PointNet to learn point features for correspondence. {DeepMapping \cite{ding2019deepmapping} uses a deep network to register multiple point clouds to a global reference frame.} 3DSmoothNet \cite{gojcic2019perfect} uses Gaussian
smoothing to voxelize points in the neighborhood, followed by a Siamese deep
network to learn local point descriptors. 3DFeat-Net
\cite{yew20183dfeat} uses weak supervision to learn correspondences from
GPS/INS tagged point clouds. UnsupervisedR\&R \cite{el2021unsupervisedr}
in an unsupervised method that uses differentiable alignment and
rendering. These methods are {usually} coupled with {random sample consensus} (RANSAC)
\cite{fischler1981random} for robust geometric registration. Deep learning
methods can yield 3D point descriptors as a byproduct.  Yet, they are
mainly optimized for a single task, {such as}, classification, segmentation or
registration. {Furthermore, they tend} not to expand the point neighborhood
successively. 

\subsection{Successive Subspace Learning (SSL)}\label{subsec:ssl}

The successive subspace learning (SSL) paradigm was introduced for point
cloud classification (called PointHop) in \cite{zhang2020pointhop} and
for image classification (called PixelHop) in \cite{chen2020pixelhop},
respectively. The idea was originated from the Saab (successive
approximation with adjusted bias) transform, which is a variant of
principal component analysis (PCA), in \cite{kuo2019interpretable}.  The
Saab transform adds a bias term to the PCA transform to address the sign
confusion problem when multiple PCA stages are in cascade. 

PointHop uses {the} statistics of 3D points to learn point cloud features in
an unsupervised one-pass manner. This procedure is summarized {as follows. First},
the attributes of a local point are constructed based on the distribution of points in its local neighborhood. All point attributes from the
training data are collected and their covariance matrix is analyzed to
define the Saab transform at the first PointHop unit. This process is
repeated, which leads to multiple PointHop units. The corresponding
receptive field grows as the hop number increases. Later, the
channel-wise Saab (c/w Saab) transform was introduced in PointHop++
\cite{zhang2020pointhop++}. The c/w Saab transform is more effective
than the Saab transform {with regard to} computational complexity and storage
complexity (i.e., model size).  Features at different hop units
of PointHop (or PointHop++) are pooled to {obtain} the global feature vector
and fed to a classifier for the classification task. Furthermore, UFF
\cite{zhang2020unsupervised} extended this framework for point cloud
part-segmentation.  PointHop and PointHop++ consist of two modules: 1)
unsupervised feature extraction and 2) supervised learning for
classification. The proposed R-PointHop method {leverages} the first
module for the registration task. 

Another closely related work is the salient points analysis (SPA) method
\cite{kadam2020unsupervised}. It is an unsupervised point cloud
registration method. SPA selects a set of salient points based on the
PCA in local neighborhood of points and uses PointHop++ to learn point
features and build correspondence among salient points for
transformation estimation. However, SPA ignores the long-range
neighborhood information in the salient point selection process.  An
inconsistent choice of salient points may lead to incorrect
registration. Also, selected salient points may not provide a clue on
which points to match when there is only a partial overlap between the two
point clouds. {In R-PointHop,} we address these shortcomings {by} using both short- and
long-range features to decide proper correspondences.

\begin{figure*}[htbp]
\centerline{\includegraphics[width=7in]{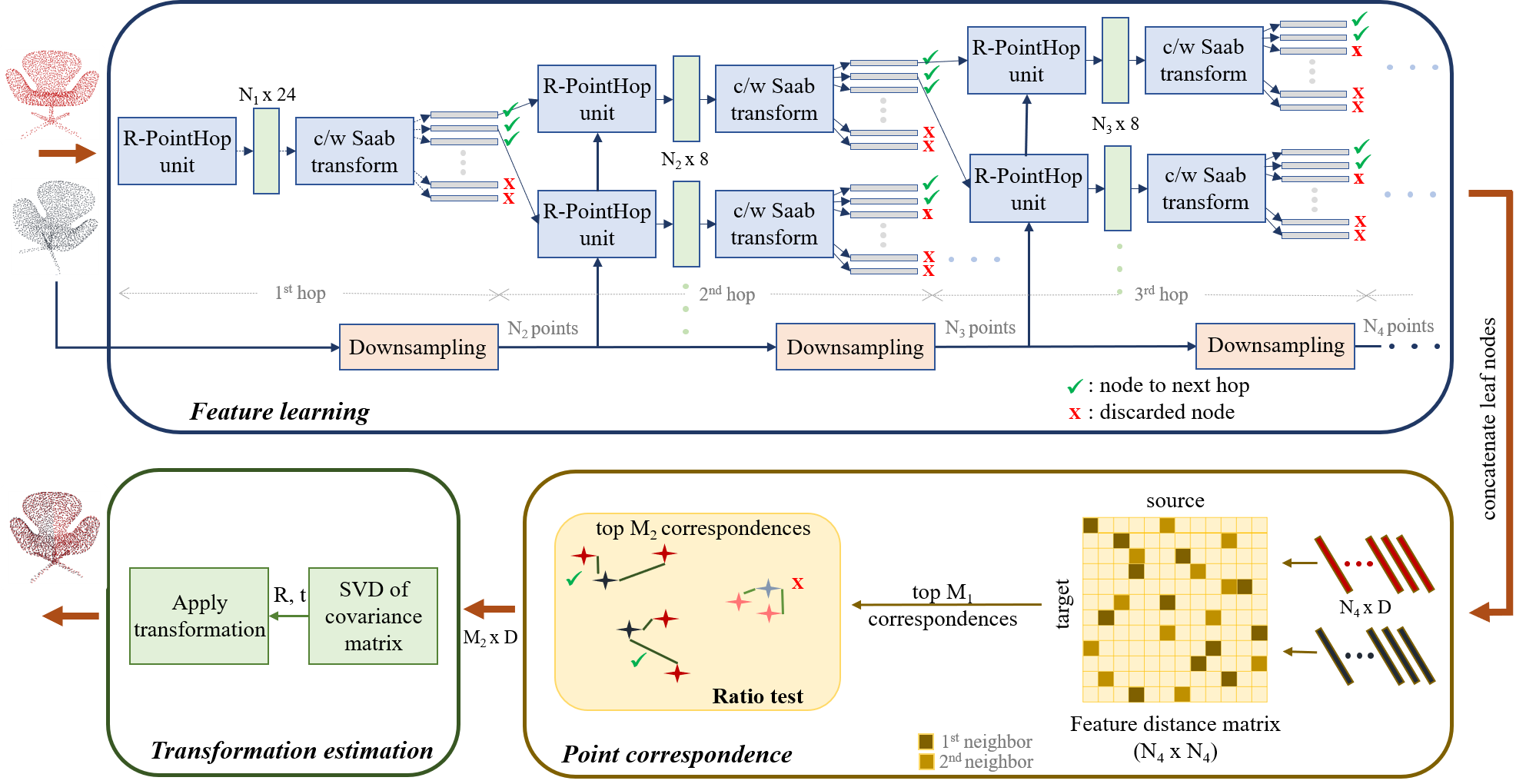}}
\caption{The system diagram of the proposed R-PointHop method, which
consists of three modules: 1) feature learning, 2) point correspondence,
and 3) transformation estimation.}\label{fig:blockdiagram}
\end{figure*}

\subsection{Rotation-Invariant Features}\label{subsec:rifl}

Rotation-invariant features do not change when the point cloud undergoes
{any} external rotation.  {These features} are desirable for robust 3D
correspondence. They are also useful for other tasks (e.g.,
classification and segmentation) in presence of different viewpoints.
Several earlier methods used the local reference frame (LRF) to design
3D descriptors that are invariant under rotation. The LRF idea lies in
the adoption of properties (e.g., distances, angles, principal
components, etc.) that are preserved under rigid transformations.
Comparison of several LRF designs is given in \cite{yang2019evaluating}.
Modern learning-based methods handle rotation invariance in different
ways. A naive approach is to augment the training data by rotating point
clouds by an arbitrary amount. Although it helps a model learn from
samples {with} different rotations during training, it does not guarantee
rotation invariance explicitly. Other methods bring point clouds to a
canonical frame before further processing. There exist separate networks
for pre-alignment. The spatial transformer network (STN)
\cite{jaderberg2015spatial} can align images to a canonical form. The
T-Net module in PointNet is another example that predicts a
transformation to align a point cloud before feature learning.
IT-Net \cite{yuan2018iterative} aligns point clouds to a canonical form
using an iterative network. The plane of symmetry in objects is detected
in \cite{nagar2019detecting}. It gives three axes {which} represent the 3D object in canonical form. Another approach is to design a convolution
operator that is invariant under rotation \cite{zhang2019rotation},
\cite{rao2019spherical}. The PPF-FoldNet \cite{deng2018ppf} learns
rotation-invariant features for point correspondence. 

PointHop and PointHop++ {both} use pre-aligned point clouds to learn features
which are not rotation-invariant.  SPA \cite{kadam2020unsupervised}
fails in registration when {the} rotation angles are larger, {because} it is
derived from PointHop++ and does not take rotation-invariance into {consideration}. Similarly, PointHop and PointHop++ do not perform well in the
classification task if an object is not pre-aligned. Here, we solve this
alignment problem by learning rotation-invariant features in an
unsupervised manner. We will show in Sec. \ref{sec:experiments} that
R-PointHop outperforms SPA by a large margin. It also makes PointHop and
PointHop++ more robust {with regard to} point cloud classification {becuase} it can pre-align 3D point clouds to a canonical form.

\section{Proposed R-PointHop Method}\label{sec:method}

The point cloud registration problem is to find a rigid transformation
(including rotation and translation) that optimally aligns two
point clouds, where one is the target point cloud denoted by $F \in
{\mathbb R}^3$ and the other is the source point cloud denoted by $G \in
{\mathbb R}^3$.  The source is obtained by applying an unknown rotation
and translation to the target.  The rotation can be expressed in form of a
rotation matrix, $R \in SO(3)$, where $SO(3)$ is a special orthogonal
group ({\em i.e.} a 3D rotation group in the Euclidean space). The
translation vector, $t \in {\mathbb R}^3$, defines the same displacement
vector for all points in the 3D space. Given $F$ and $G$, the goal is to
find an optimal $R^* \in SO(3)$ and translation $t^* \in {\mathbb R}^3$
that minimize the mean squared error between matching points given by
\begin{equation}\label{eq:registration}
E(R,t)=\frac{1}{N}\sum\limits_{i=0}^{N-1}\|R^*{\bf f}_i+t^*-{\bf g}_i\|^2,
\end{equation}
where $({\bf f}_i,{\bf g}_i)$ denotes a pair of $N$ selected matching points.
Although the actual number of points in each point could be larger than
$N$, the error is defined over the $N$ matching points for convenience.
The system diagram of the proposed R-PointHop method is shown in Fig.
\ref{fig:blockdiagram}. It contains three main modules: 1) feature
learning, 2) point correspondence, and 3) transformation estimation. 
{These modules} are detailed below.

\begin{figure}[htbp]
\centerline{\includegraphics[width=3.2in]{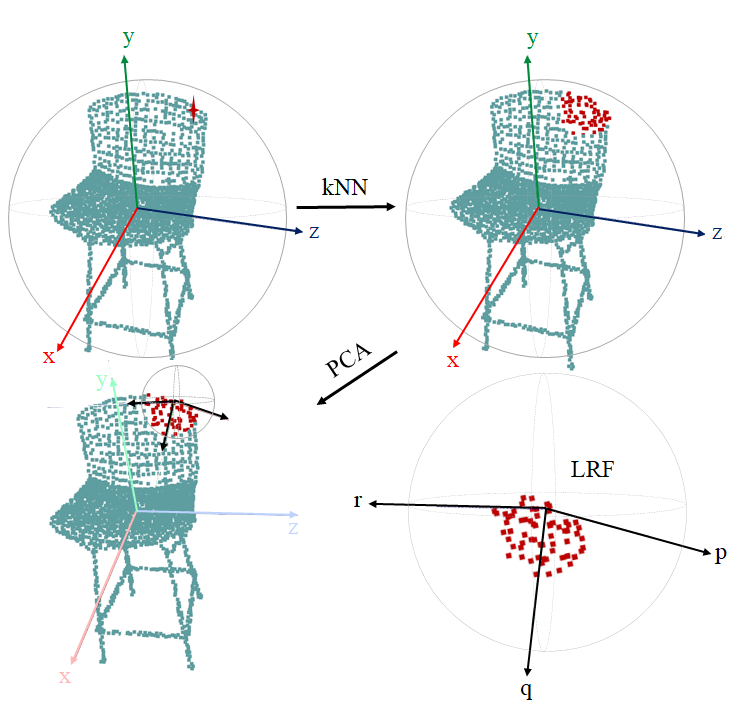}}
\caption{Illustration of the local reference frame (LRF).  The top-left
sub figure shows the query point (marked in red) and the XYZ coordinate
axes.  The top-right sub figure shows the nearest $K$ neighbors of the
query point (marked in red).  The bottom-left sub figure shows the three
eigenvectors of the local PCA of the 3D coordinates of points in the
marked neighborhood. The bottom right sub figure shows the LRF of the
query point.}\label{fig:lrf}
\end{figure}

\begin{figure*}[htbp]
\centerline{\includegraphics[width=7in]{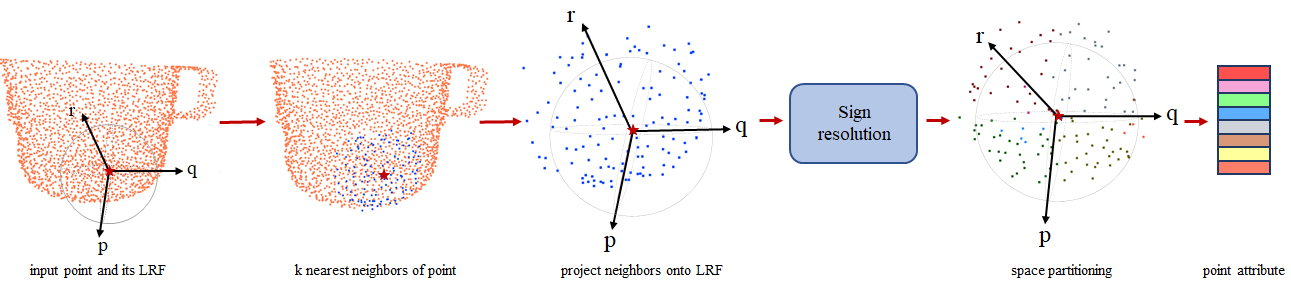}}
\caption{Illustration of point attribute construction. First, the
$K$ nearest neighbors of a target point are fetched, and they are
projected to its LRF. Next, the sign ambiguity is resolved using a
proper reflection matrix. Afterwards, the LRF is partitioned into eight
octants and the mean of points in every octant is calculated. Finally,
the mean coordinates are concatenated to get a 24-D attribute
vector.}\label{fig:attribute}
\end{figure*}

\subsection{Feature Learning}\label{subsec:module_1}

In the feature extraction process, a
$D$-dimensional feature vector {is learned} for every point in a hierarchical manner.
The feature learning function, $g(\cdot)$, takes input points
of dimension $D_0$ and outputs points with feature dimension $D$. Here,
$D_0$ represents 3D point coordinates along with optional point
properties like the surface normal, color, etc. Stage $h$ in the
hierarchical feature learning process (or $h$-th hop) is a function
$g_h(\cdot)$ that takes the point feature of the previous hop of
dimension $D_{h-1}$ and outputs feature of dimension $D_h$.

To find feature $f_{i,h}$ of the $i$-th point in
the $h$-th hop, the input to $g_h(\cdot)$ includes point coordinates $x_i$,
features of the $i$-th point from the previous hop $f_{i,h-1}$,
coordinates of $K$ neighboring points $x_j$ in hop $h$, features of
neighboring points $f_{j,h-1}$ from previous hop, and a reference frame
$F$. Thus, $f_{i,h}$ is given by
\begin{equation}
    f_{i,h} = g_h(x_i,x_j, f_{i,h-1}, f_{j,h-1}, F)
\end{equation}

There are several choices of $g_h(\cdot)$, which can
be determined based on whether the goal is to learn a local or global
feature. For R-PointHop, we choose $g_h(\cdot)$ such that
\begin{equation}\label{fh_R-PointHop}
    f_{i,h} = g_h(x_j-x_i, f_{i,h-1}, f_{j,h-1}, LRF(x_i,x_j)),
\end{equation}
where $LRF(x_i,x_j)$ is the local reference frame
centered at $x_i$ ({see Sec. \ref{subsec:LRF} below}). This choice
of $g_h(\cdot)$ encodes only the local patch information and loses the
global shape structure. In contrast, the PointHop and SPA learning
functions are given by
\begin{equation}\label{fh_SPA}
    f_{i,h} = g_h(x_j, f_{i,h-1}, f_{j,h-1}, XYZ),
\end{equation}
where $XYZ$ denotes that points are always expressed
in the original frame of reference. Although this learning function
captures the global shape structure as the spatial locations of the
neighborhood patches $x_j$ are preserved, it limits the registration
performance {in presence of} a large rotation angle.

Instead, R-PointHop {keeps the local
position information with LRF.} This is desired for
registration, since matching points (or patches), which could be
spatially far apart, are still close in the feature space now. In contrast,
the global position information is vital for the classification task
since we are interested in how different local patches connect to other
patches that form the overall shape. Thus, the use of the global
coordinates in the classification problem is justified.

\subsubsection{Local Reference Frame (LRF)} \label{subsec:LRF}

The Principal Components Analysis (PCA) of the 3D coordinates of points
in a local neighborhood {provides insight into} the local surface structure.
The third eigenvector of the PCA can be taken as a rough estimate of the
surface normal. Although the local PCA computation was used in SPA
\cite{kadam2020unsupervised} to select salient points, SPA pays more
attention to the eigenvalue rather than the eigenvector. Since the local
PCA centered at a point is invariant under a rotation of the point
cloud, the local PCA of true corresponding points should be similar.
This observation serves as the basis to derive the local reference
frame (LRF) for every point. That is, we consider $K$ nearest neighbors
of a point and conduct the PCA on their 3D coordinates.  This results in
three mutually orthogonal eigenvectors. They are sorted in a decreasing
order of the associated variances (or eigenvalues).  We use $X$, $Y$,
$Z$ as a convention to represent the original reference in which the
point clouds are defined.  For the LRF, we use $P, \ Q, \ R$ to label
the three axes corresponding to the three eigenvectors of largest,
middle, and smallest eigenvalues. The eigenvectors come with a sign
ambiguity {problem} since the negative of an eigenvector is still an eigenvector.
There are various methods to tackle the sign ambiguity problem. The
distribution of neighboring points at every hop is exploited in our
work to handle this ambiguity and is be discussed later. Then, we can define positive
eigenvectors $(p^+, \ q^+, \ r^+)$ and negative eigenvectors $(p^-, \
q^-, \ r^-)$ for each point. They are unique and serve as the LRF for
every point. An example is illustrated in Fig. \ref{fig:lrf}. 

\subsubsection{{Constructing Point Attributes}}
To construct the attributes of a target point, we find its $K$ nearest
neighbors. They can be the same as those in the previous step or in a
larger neighborhood depending on the point density and the total number
of points.  For each point in the neighborhood, we transform its XYZ
coordinates to the LRF of the target point. The eigenvectors $(p^+, \
q^+, \ r^+)$ are used as default axes. To address the sign ambiguity of
each axis individually, we consider the 1D coordinates of $K$ points of an
axis, find the median point and calculate the first-order moment about
the median point. Initially, we can assign $p^+$ or $p^-$ arbitrarily.
The first-order left and right moments are given, respectively, by
\begin{eqnarray}\label{eq:axis}
M_p^l & = & \sum_i |p_i-p_m| \quad \forall \ p_i < p_m, \\
M_p^r & = & \sum_i |p_i-p_m| \quad \forall \ p_i > p_m,
\end{eqnarray}
where $p_i$ is the 1D coordinates of point $i$ projected to $p^+$ and
$p_m$ is the projected value of the median point. If $M_p^r > M_p^l$, we
{retain} original assignment {of} $p^+$/$p^-$. Otherwise, we swap the
assignment to ensure the direction with {the} larger first-order moment is the positive axis. This can be implemented {by} post-multiplying the
local data matrix of dimension $K \times 3$ with a diagonal reflection
matrix, $R' \in R^{3 \times 3}$, whose diagonal elements are either $1$
or $-1$ depending on the chosen sign. That is,
\begin{equation}
    R'_{ii} =
\begin{cases}
    1,& \text{if } M_i^l < M_i^r, \\
    -1,& \text{otherwise},
\end{cases}
\end{equation}
and
\begin{equation}
    R'_{ij} = 0, \quad \text{if } i \neq j.
\end{equation}
{Here,} we deliberately use $R'$ for the reflection matrix in above so as to
avoid confusion with {the} rotation matrix $R$.  The 3D space of {the} $K$ nearest
neighbor points is partitioned into eight octants {of} the LRF, centered at
the target point. For each octant, we calculate the mean of {all} the 3D
coordinates of points in that octant and concatenate all eight means to
get a 24D vector, which is the attributes of the target point. The
same process is conducted on all points in the point cloud. The octant
partitioning and grouping is similar to that of PointHop, {but} the difference
lies in the use of the LRF in R-PointHop. The attribute construction 
process is illustrated in Fig. \ref{fig:attribute}. 

\begin{figure*}[htbp]
\centerline{\includegraphics[width=7in]{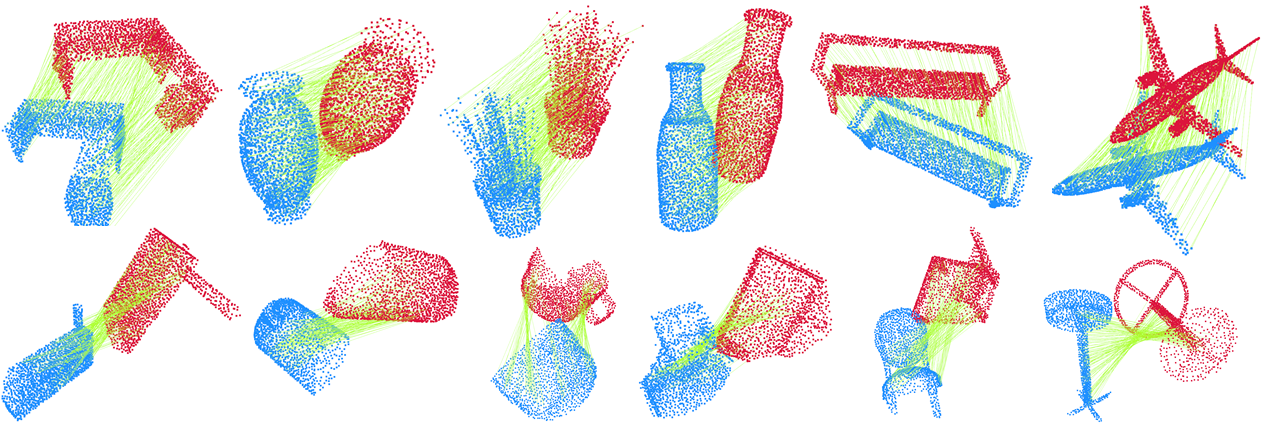}}
\caption{Correspondences found using R-PointHop, where the source point cloud
is shown in red and the target is shown in blue.}\label{fig:correspondences}
\end{figure*}

\subsubsection{Multi-hop Features}

The 24D attributes of all points of point clouds from the training set are
collected, and the Saab transform \cite{zhang2020pointhop++} is
conducted to {obtain a} 24D spectral representation. This is the output of
hop \#1.  We compute the energy of each node as done in
\cite{zhang2020pointhop++} and pass the nodes of energy greater than
threshold $T$ to the next hop and discard the nodes of energy smaller
than threshold $T$. In PointHop++, the nodes with energy less than
threshold $T$ are collected as leaf nodes. Here, we discard them to
avoid mismatched correspondences. This is because hop \#1 features carry
more local structure information which may be similar in different
regions of the point cloud.  Proceeding to the next hop, the point cloud
is downsampled using the Farthest Point Sampling (FPS). FPS
ensures that the structure of the point cloud is preserved after
downsampling. It also helps reduce computations and grow the receptive
field quickly.  At hop \#2, the attribute construction process is
repeated at every node passed on from hop \#1. Since these features are
uncorrelated, we can handle them separately and apply the channel-wise
Saab transform \cite{zhang2020pointhop++} starting from hop \#2 and
beyond. Each dimension is treated as a node in the feature tree in the
channel-wise Saab transform. The $K$ nearest neighbors of a target point
at hop \#2 are found, which are different from hop \#1 neighbors due to
the downsampling operation. They are represented using the LRF of the
target point found in the first step. Since the set of $K$ {nearest neighbor} points {has}
changed, we have to decide the appropriate sign again. The LRF is
partitioned into eight octants, in each of which we take the mean of the
1D feature of all points in that octant.  The eight means are
concatenated to get 8D hop \#2 attributes for a node. All the point
attributes are collected and the channel-wise Saab transform is used to
get the 8D spectral representation. This process is repeated for all
nodes at hop \#2. The multi-hop learning process continues for four
hops. All 1D spectral components at the end of hop \#4 are concatenated
the get the feature vector of a point.  The final feature dimension
depends on the choice of different parameters including the neighborhood
size, number of points to be downsampled at every hop, and the energy
threshold for channel-wise Saab transform.  These parameters can be
different at different hops. A set of model parameters will be presented
in Sec. \ref{sec:experiments}. 

The rotation/translation {invariance} property of
R-PointHop comes from the use of the Local Reference Frame (LRF). {The} LRF is
derived by applying PCA to points in a local neighborhood. In the
attribute building step, we collect points in a local neighborhood and
project them onto the local coordinate system. This ensures that when
the point cloud undergoes any rotation or translation, the coordinates
of neighboring points remain the same since the LRF also rotates and
translates accordingly. In subsequent stages, we keep projecting the
neighboring points onto the LRF to ensure the rotation/translation
{invariance} property is preserved at every stage.

\subsection{Establishing Point Correspondences}\label{subsec:module_2}

The trained R-PointHop model is used to extract features from the target
and the source point clouds. A feature distance matrix is calculated
whose $ij^{th}$ element is the $l_2$ distance between the feature of the
$i^{th}$ point in the target and the $j^{th}$ point in the source.  The
minimum value along the $i^{th}$ row gives the point in the source which
is closest to the $i^{th}$ point in the target in the feature space.
These pairs of points nearest in the feature space are used as an
initial set of correspondences. {Next, we select a subset} of good
correspondences. To do so, the correspondences are first ordered in the
increasing $l_2$ distance between features of matching points. Top $M_1$
correspondences are selected using this criterion. We use the ratio test
to further select a smaller set of $M_2$ correspondences. That is, the
distance to the second nearest neighbor is found as the second minima
along the row in the distance matrix. The ratio between the distance to
the first neighbor and that to the second neighbor is calculated. A
smaller ratio indicates a higher confidence of match.  Top $M_2$
correspondences are selected using the ratio test. These points are used
to find the rotation and translation. Instead of choosing $M_1$ and
$M_2$ points explicitly, we can alternatively set two thresholds $t_1$
and $t_2$, where $t_1$ is for the minimum $l_2$ distance between
matching features and $t_2$ for the minimum ratio. These
hyper-parameters are selected empirically in our experiments.  It is
worthwhile to comment that SPA \cite{kadam2020unsupervised} presented an
analogous method to select a subset of correspondences.  The main
difference between R-PointHop and SPA lies in the fact that SPA uses local PCA
only to find salient points in the point cloud. It ignores the rich
multi-hop spectral information. In contrast, R-PointHop uses multi-hop
features to select a high-quality subset of correspondences. 

\begin{figure*}[htbp]
\centerline{\includegraphics[width=7in]{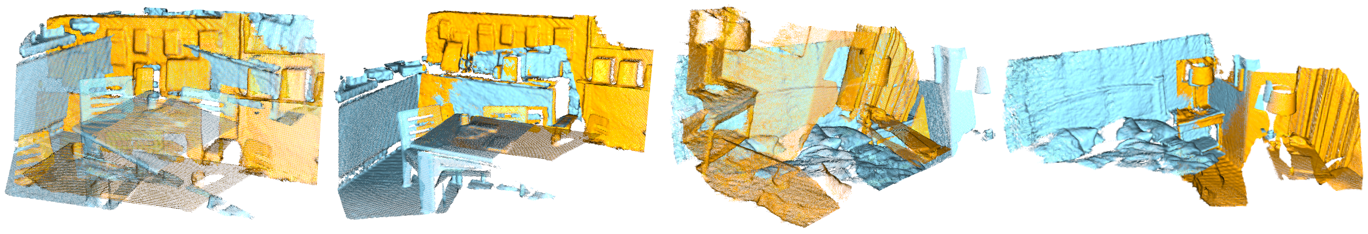}}
\caption{Registration of indoor point clouds from
3DMatch dataset: point clouds from 7-Scenes (the left two) and point clouds
from SUN3D (the right two))}.\label{fig:3dmatch}
\end{figure*}

\subsection{Estimating Transformation}\label{subsec:module_3}

The ordered pairs of corresponding points $({\bf f}_i,{\bf g}_i)$ are
used to estimate the optimal rotation $R^*$ and translation $t^*$ {that}
minimizes the error function as given in Eq.  (\ref{eq:registration}). A
closed-form solution to this optimization problem was given in
\cite{schonemann1966generalized}. It can be solved numerically using the
singular value decomposition (SVD) of the data covariance matrix.  The
procedure is summarized below. 
\begin{enumerate}
\item Find the mean point coordinates from the correspondences by
\begin{equation}
\Bar{{\bf f}}=\frac{1}{N}\sum\limits_{i=0}^{N-1} {\bf f}_i, \quad 
\Bar{{\bf g}}=\frac{1}{N}\sum\limits_{i=0}^{N-1} {\bf g}_i.
\end{equation}
Then, {compute the covariance matrix}
\begin{equation}
Cov(F,G)=\sum\limits_{i=0}^{N-1}({\bf f}_i-\Bar{{\bf f}})({\bf g}_i-\Bar{\bf g})^T.
\end{equation}
\item Conduct SVD on the covariance matrix
\begin{equation}
Cov(F,G)=USV^T,
\end{equation}
where $U$ is the matrix of left singular vectors, $S$ is the diagonal matrix
containing singular values and $V$ is the matrix of right singular
vectors. In this case, $U, \ S$, {and} $V$ are $3 \times 3$ matrices. 
\item The optimal rotation matrix $R^*$ is given by
\begin{equation}
R^*=V U^T.
\end{equation}
The optimal translation vector $t^*$ can be found using $R^*$ and the
means $\Bar{x}$ and $\Bar{y}$: 
\begin{equation}
t^*=-R^*\Bar{{\bf f}}+\Bar{{\bf g}},
\end{equation}
$R^*$ and $t^*$ are {then} used to align the source with the target. 
\end{enumerate}
Finally, the aligned source point cloud $(G')$ is given by
\begin{equation}
G'=R^{*T}(G-t^*),
\end{equation}
where $R^{*T}$ is the transpose of $R^{*}$ which applies the inverse
transformation. Unlike SPA which iteratively aligns the source to
target, R-PointHop is not iterative and point cloud registration is 
completed in one run. 

\section{Experimental Results} \label{sec:experiments}

Experiments are performed on point clouds {of}
indoor scenes {and} 3D objects. We begin our discussion with indoor
scene registration.

\subsection{Indoor Scene Registration}

We {trained} and evaluated R-PointHop on indoor point
cloud scans from the 3DMatch dataset \cite{zeng20173dmatch}. This
dataset is an ensemble of several RGB-D reconstruction datasets such as
7-Scenes \cite{shotton2013scene} and SUN3D \cite{xiao2013sun3d}. The
dataset comprises of various indoor scenes {such as} bedroom, kitchen,
office, lab, and hotel. There are 62 scenes in total, which are split
into 54 training {scenes} and 8 testing {scenes}. {Each} scene is further divided
into various partial overlapping point clouds consisting of 200-700K
points.

During training and evaluations, 2,048 points are
randomly sampled from each point cloud scan. By inspecting several
examples visually, randomly selected points roughly span the entire set
and, hence, computationally intensive sampling schemes such as FPS can
be avoided. 256 neighboring points are used to determine the LRF. We
append point coordinates with the surface normal and geometric features
(e.g., linearity, planarity, sphericity, eigen entropy, etc.
\cite{hackel2016fast}) obtained from eigenvalues of local PCA as point
attributes. Since local PCA is already performed in the LRF computation,
their eigenvalues are readily available. Furthermore, we use RANSAC
\cite{fischler1981random} to estimate the transformation. Some
successful registration results are shown in Fig. \ref{fig:3dmatch}.

We compare R-PointHop with 3DMatch
\cite{zeng20173dmatch} and PPFNet \cite{deng2018ppfnet} since they are
among early supervised deep learning methods developed for indoor scene
registration. Furthermore, several model-free methods such as SHOT
\cite{tombari2010unique}, Spin Images \cite{johnson1997spin} and FPFH
\cite{rusu2009fast} are also included for performance benchmarking. All
methods are evaluated based on 2048 sampled points for fair comparison.
By following the evaluation method given by \cite{Choi_2015_CVPR}, we
report the average recall and precision on the test set. The results
are summarized in Table \ref{tab:3DMatch}. As shown in the table,
R-PointHop offers the highest recall and precision. It outperforms
model-free methods by a significant margin. Its performance is slightly
superior to {that of} PPFNet.

\begin{table}[htbp]
\centering
\caption{Registration performance comparison on the 3DMatch dataset.} \label{tab:3DMatch}
\renewcommand\arraystretch{1.3}
\newcommand{\tabincell}[2]{\begin{tabular}{@{}#1@{}}#2\end{tabular}}
\begin{tabular}{c c c} \hline 
{\bf Method} & \tabincell{c}{{\bf Recall}}  &  \tabincell{c}{{\bf Precision}}  \\ \hline 
SHOT \cite{tombari2010unique}  & 0.27 & 0.17   \\ \hline
Spin Images \cite{johnson1997spin}  & 0.34 & {0.18}  \\ \hline
{FPFH \cite{rusu2009fast}}  & {0.41} & {0.21}   \\ \hline
{3DMatch \cite{zeng20173dmatch}}  & {0.63}  & {0.24}   \\ \hline
{PPFNet \cite{deng2018ppfnet}}   & {0.71} & {\bf 0.26}  \\ \hline
{R-PointHop}  & {\bf 0.72} & {\bf 0.26}  \\ \hline
\end{tabular}
\end{table}

\begin{figure*}[htbp]
\centerline{\includegraphics[width=7in]{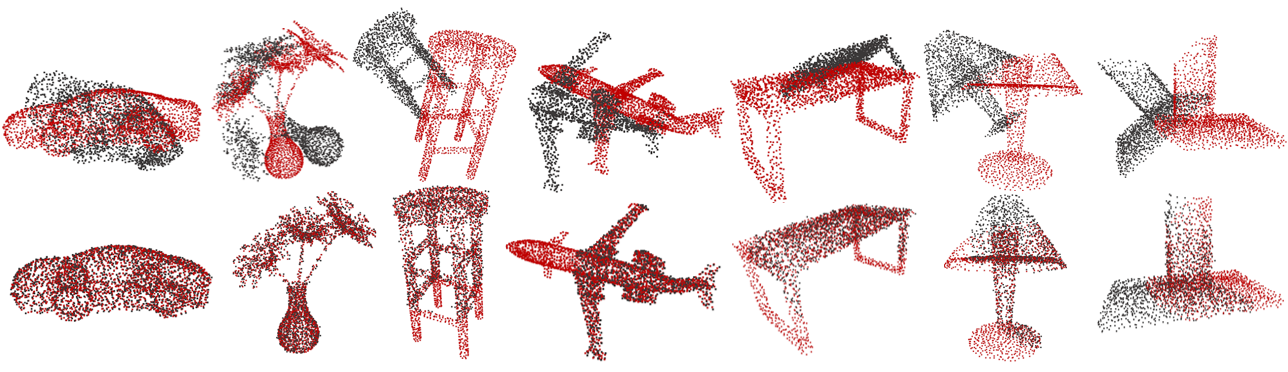}}
\caption{Registration of seven point clouds from the ModelNet40 dataset
using R-PointHop. The first row shows source point clouds (in black) and
their target point clouds (in red), respectively. The second row shows
registration results.  Both the source and the target are complete point
clouds in columns \#1 and \#2.  The source in columns \#3-\#5 contains
only part of the target. Both the source and the target are partial in
columns \#6 and \#7.}\label{fig:examples}
\end{figure*}

\subsection{Object Registration}

{Next}, we trained R-PointHop on the ModelNet40 dataset \cite{wu20153d}. It is a
synthetic dataset consisting of $40$ categories of CAD models of common
objects such as car, chair, table, airplane, {and} person. It comprises 12,308 point cloud models in total, which are split into 9840
training {models} and 2468 testing {models}. Every point cloud model consists of
2,048 points. {The} point clouds are normalized to fit
within a unit sphere. For the task of 3D registration, we follow the
same experimental setup as DCP \cite{wang2019deep} and PR-Net
\cite{wang2019prnet} for fairness. 

The following set of parameters are chosen as the default of R-PointHop
for object registration.
\begin{itemize}
    \item Number of initial points: 1,024 points (randomly sampled from the 
    original 2,048 points)
    \item Point Attributes: point coordinates only\footnote{Although the surface normal and 
          geometric features were included for indoor registration, they are removed in the 
          context of object registration.}
    \item Neighborhood size for finding LRF: 64 nearest neighbors
    \item Number of points in each hop: $1024, \ 768, \ 512, \ 384$
    \item Neighborhood size in each hop: $64, \ 32, \ 48, \ 48$
    \item Energy threshold: 0.001
    \item Number of top correspondences selected: 256
    \item Number of correspondences selected after the ratio test: 128
\end{itemize}

We compare R-PointHop with the following six methods:
\begin{itemize}
\item three model-free methods \\
ICP \cite{besl1992method}, Go-ICP \cite{yang2015go} and FGR \cite{zhou2016fast}.
\item two supervised-learning-based methods \\
PointNetLK \cite{aoki2019pointnetlk} and DCP \cite{wang2019deep}.
\item one unsupervised-learning-based method \\
SPA \cite{kadam2020unsupervised}.
\end{itemize}
For ICP, Go-ICP and FGR we use the open-source implementation in Open3D
library \cite{Zhou2018}. 

In Secs. \ref{subsec:5.1}-\ref{subsec:5.5}, we apply a random rotation
to the target point cloud about its three coordinate axes. Each rotation
angle is uniformly sampled in $[0^{\circ},45^{\circ}]$. Then, a random
uniform translation in $[-0.5,0.5]$ is applied along the three axes to
get the source point cloud. For training, only the target point clouds
are used.  We report the Mean Square Error (MSE), the Root Mean Square
Error (RMSE) and the Mean Absolute Error (MAE) between the ground truth and the predicted rotation angles and the predicted translation vector.
In Sec. \ref{subsec:5.5}, we align real world point clouds from the
Stanford 3D scanning repository \cite{turk1994zippered},
\cite{curless1996volumetric}, \cite{krishnamurthy1996fitting}. In Sec.
\ref{subsec:5.6}, we show that R-PointHop can be used for global
registration as well as an initialization for ICP. In Sec.
\ref{subsec:5.7}, we explain the use of R-PointHop as a general 3D point
descriptor. 

\subsubsection{Registration on Unseen Data}\label{subsec:5.1}

In this experiment, we {trained} R-PointHop from training samples of all 40
classes. For {evaluation, registration was performed on} point clouds from the test data are used. The
results are reported in Table \ref{tab:Result1}. We see that R-PointHop
clearly outperforms all six benchmarking methods.  Two {sets of} target and source
point clouds and their registered results are shown in the first two
columns of Fig. \ref{fig:examples}. To plot point clouds, we use the
Open3D library \cite{Zhou2018}. 

\begin{table}[htbp]
\centering
\caption{Registration on unseen point clouds} \label{tab:Result1}
\renewcommand\arraystretch{1.3}
\newcommand{\tabincell}[2]{\begin{tabular}{@{}#1@{}}#2\end{tabular}}
\resizebox{\columnwidth}{!}{
\begin{tabular}{c c c c c c c} \hline 
\bf Method & \tabincell{c}{\bf MSE\\(R)}  &  \tabincell{c}{\bf RMSE\\(R)}  
& \tabincell{c}{\bf MAE\\(R)} & \tabincell{c}{\bf MSE\\(t)}  &  \tabincell{c}{\bf RMSE\\(t)} 
& \tabincell{c}{\bf MAE\\(t)} \\ \hline 
\tabincell{c} 
ICP \cite{besl1992method}  & 451.11 & 21.24  & 17.69 & 0.049701 & 0.222937 & 0.184111 \\ \hline
Go-ICP \cite{yang2015go}  & 140.47 & 11.85  & 2.59 & 0.00659 & 0.025665 & 0.007092 \\ \hline
FGR \cite{zhou2016fast}  & 87.66 &9.36  & 1.99 & 0.000194 & 0.013939 & 0.002839 \\ \hline
PointNetLK \cite{aoki2019pointnetlk}  & 227.87 & 15.09  & 4.23 & 0.000487 & 0.022065 & 0.005405 \\ \hline
DCP \cite{wang2019deep}   & 1.31 & 1.14  & 0.77 & 0.000003 & 0.001786 & 0.001195  \\ \hline
SPA \cite{kadam2020unsupervised}  & 318.41 & 17.84 & 5.43 & 0.000022 & 0.004690 & 0.003261 \\ \hline
R-PointHop  & \bf{0.12} & \bf{0.34} & \bf{0.24} & \bf{0.000000} & \bf{0.000374} & \bf{0.000295}  \\ \hline
\end{tabular}}
\end{table}

\begin{figure*}[htbp]
\centerline{\includegraphics[width=6.8in,height=2.5in]{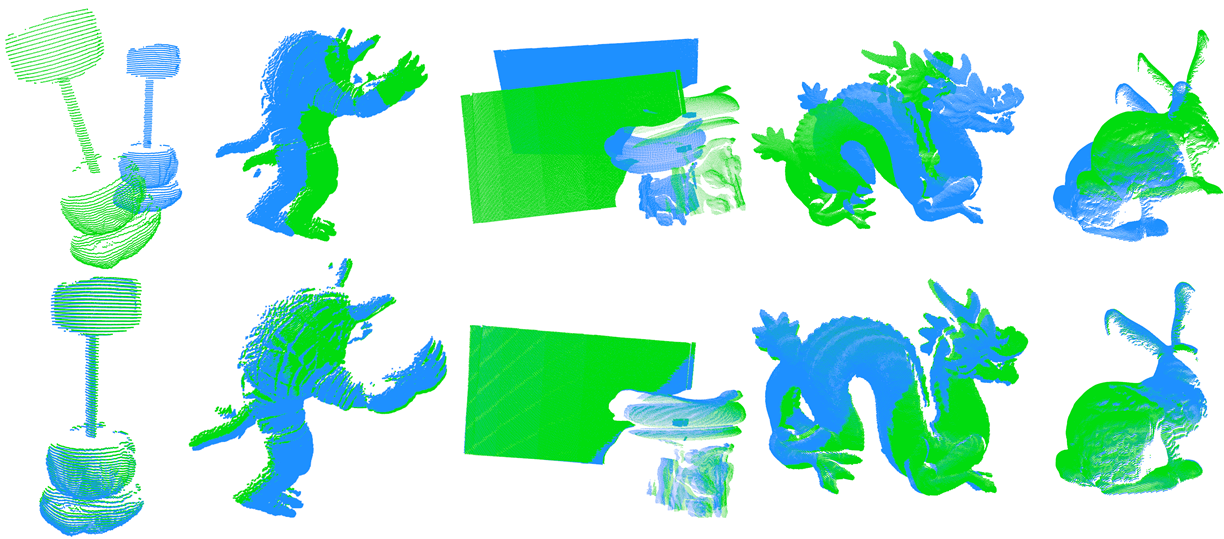}}
\caption{Registration of point clouds from the Stanford 3D scanning
repository, where the objects are (from left to right): drill bit,
armadillo, Buddha, dragon and bunny. The top row shows input point
clouds while the bottom row shows the registered output.} \label{fig:Stanford3D}
\end{figure*}

\subsubsection{Registration on Unseen Classes}\label{subsec:5.2}

We derive R-PointHop only from the first 20 classes of the ModelNet40
dataset. For registration, test samples from the remaining 20 classes
are used. As shown in Table \ref{tab:Result2}, R-PointHop can generalize
well on unseen classes. PointNetLK and DCP have relatively larger errors
as compared to their errors in Table \ref{tab:Result1}.  This indicates
that the use of object labels makes these methods biased to the seen
categories. For the first three methods, the results are comparable with
those in Table \ref{tab:Result1} as there is no training involved.  For
SPA and R-PointHop, their errors are similar to those of unseen object
classes. {This demonstrates} the advantage of unsupervised learning {methods for} registration
of unseen classes. 

\begin{table}[htbp]
\centering
\caption{Registration on unseen classes} \label{tab:Result2}
\renewcommand\arraystretch{1.3}
\newcommand{\tabincell}[2]{\begin{tabular}{@{}#1@{}}#2\end{tabular}}
\resizebox{\columnwidth}{!}{
\begin{tabular}{c c c c c c c} \hline 
\bf Method & \tabincell{c}{\bf MSE\\(R)}  &  \tabincell{c}{\bf RMSE\\(R)}  
& \tabincell{c}{\bf MAE\\(R)} & \tabincell{c}{\bf MSE\\(t)}  &  \tabincell{c}{\bf RMSE\\(t)} 
& \tabincell{c}{\bf MAE\\(t)} \\ \hline 
\tabincell{c} 
ICP \cite{besl1992method}  & 467.37 & 21.62  & 17.87 & 0.049722 & 0.222831 & 0.186243 \\ \hline
Go-ICP \cite{yang2015go}  & 192.25 & 13.86  & 2.91 & 0.000491 & 0.022154 & 0.006219 \\ \hline
FGR \cite{zhou2016fast}  & 97.00 & 9.84  & 1.44 & 0.000182 & 0.013503 & 0.002231 \\ \hline
PointNetLK \cite{aoki2019pointnetlk}  & 306.32 & 17.50  & 5.28 & 0.000784 & 0.028007 & 0.007203 \\ \hline
DCP \cite{wang2019deep}   & 9.92 & 3.15  & 2.01 & 0.000025 & 0.005039 & 0.003703 \\ \hline
SPA \cite{kadam2020unsupervised}  & 354.57 & 18.83 & 6.97 & 0.000026 & 0.005120 & 0.004211  \\ \hline
R-PointHop  & \bf{0.12} & \bf{0.34} & \bf{0.25} & \bf{0.000000} & \bf{0.000387} & \bf{0.000298}  \\ \hline
\end{tabular}}
\end{table}

\subsubsection{Registration on Noisy Data}\label{subsec:5.3}

In this experiment, we {were} interested in aligning a noisy source point
cloud with a target that is free from noise. A Gaussian noise with zero
mean and standard deviation of 0.01 {was} added to the source. The
registration results are presented in Table \ref{tab:Result3}. {The results demonstrate}
that R-PointHop is robust to Gaussian noise. A fine alignment step using
ICP can further reduce the error. In other words, R-PointHop can act as
a coarse alignment method in presence of noise. 

\begin{table}[htbp]
\centering
\caption{Registration on noisy point clouds} \label{tab:Result3}
\renewcommand\arraystretch{1.3}
\newcommand{\tabincell}[2]{\begin{tabular}{@{}#1@{}}#2\end{tabular}}
\resizebox{\columnwidth}{!}{
\begin{tabular}{c c c c c c c} \hline 
\bf Method & \tabincell{c}{\bf MSE\\(R)}  &  \tabincell{c}{\bf RMSE\\(R)}  
& \tabincell{c}{\bf MAE\\(R)} & \tabincell{c}{\bf MSE\\(t)}  &  \tabincell{c}{\bf RMSE\\(t)} 
& \tabincell{c}{\bf MAE\\(t)} \\ \hline 
\tabincell{c} 
ICP \cite{besl1992method}  & 558.38 & 23.63  & 19.12 & 0.058166 & 0.241178 & 0.206283 \\ \hline
Go-ICP \cite{yang2015go}  & 131.18 & 11.45  & 2.53 & 0.000531 & 0.023051 & 0.004192 \\ \hline
FGR \cite{zhou2016fast}  & 607.69 & 24.65 & 10.05 & 0.011876 & 0.108977 & 0.027393 \\ \hline
PointNetLK \cite{aoki2019pointnetlk}  & 256.15 & 16.00  & 4.59 & 0.000465 & 0.021558 & 0.005652 \\ \hline
DCP \cite{wang2019deep}   & 1.17 & 1.08  & 0.74 & 0.000002 & 0.001500 & 0.001053  \\ \hline
SPA \cite{kadam2020unsupervised}  & 331.73 & 18.21 & 6.28 & 0.000462 & 0.021511 & 0.004100  \\ \hline
R-PointHop  & 7.73 & 2.78 & 0.98 & 0.000001 & 0.000874 & 0.003748  \\ \hline
R-PointHop + ICP  & \bf{1.16} & \bf{1.08} & \bf{0.21} & \bf{0.000001} & \bf{0.000744} & \bf{0.001002}  \\ \hline
\end{tabular}
}
\end{table}

\begin{figure}[htbp]
\centerline{\includegraphics[width=3.4in,height=1.5in]{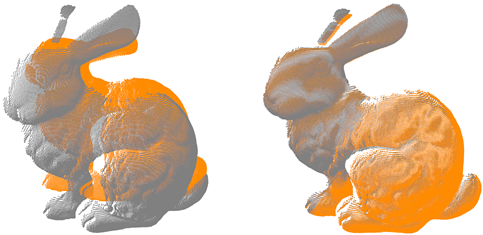}}
\caption{Registration on the Stanford Bunny dataset: the source and
the target point clouds (left) and the registered result (right).}\label{fig:Bunny}
\end{figure}

\begin{table*}[htbp]
\centering
\caption{Registration on partial point clouds (R-PointHop* indicates
choosing correspondences without the ratio test).} \label{tab:Result4}
\renewcommand\arraystretch{1.3}
\newcommand{\tabincell}[2]{\begin{tabular}{@{}#1@{}}#2\end{tabular}}
\begin{tabular}{c| c c c c c c |c c c c c c c} \hline 
\textbf{} &\multicolumn{6}{|c|}{\textbf{Registration errors on unseen objects}} 
&\multicolumn{6}{c}{\textbf{Registration errors on unseen classes}} \\ \hline
\bf Method & \tabincell{c}{\bf MSE\\(R)}  &  \tabincell{c}{\bf RMSE\\(R)}  
& \tabincell{c}{\bf MAE\\(R)} & \tabincell{c}{\bf MSE\\(t)}  &  \tabincell{c}{\bf RMSE\\(t)} 
& \tabincell{c}{\bf MAE\\(t)} & \tabincell{c}{\bf MSE\\(R)}  &  \tabincell{c}{\bf RMSE\\(R)}  
& \tabincell{c}{\bf MAE\\(R)} & \tabincell{c}{\bf MSE\\(t)}  &  \tabincell{c}{\bf RMSE\\(t)} 
& \tabincell{c}{\bf MAE\\(t)} \\ \hline 
\tabincell{c} 
ICP \cite{besl1992method}  & 1134.55 & 33.68  & 25.05 & 0.0856 & 0.2930 & 0.2500 & 1217.62 & 34.89  & 25.46 & 0.0860 & 0.293 & 0.251 \\ \hline
Go-ICP \cite{yang2015go}  & 195.99 & 13.99  & 3.17 & 0.0011 & 0.0330 & 0.0120 & 157.07 & 12.53  & 2.94 & 0.0009 & 0.031 & 0.010 \\ \hline
FGR \cite{zhou2016fast}  & 126.29 & 11.24  & 2.83 & 0.0009 & 0.0300 & 0.0080 & 98.64 & 9.93  & 1.95 & 0.0014 & 0.038 & 0.007  \\ \hline
PointNetLK \cite{aoki2019pointnetlk}  & 280.04 & 16.74  & 7.55 & 0.0020 & 0.0450 & 0.0250 & 526.40 & 22.94  & 9.66 & 0.0037 & 0.061 & 0.033 \\ \hline
DCP \cite{wang2019deep}   & 45.01 & 6.71  & 4.45 & 0.0007 & 0.0270 & 0.0200 & 95.43 & 9.77 & 6.95 & 0.0010 & 0.034 & 0.025 \\ \hline
PR-Net \cite{wang2019prnet} & 10.24 & 3.12 & 1.45 & 0.0003 & 0.0160 & 0.0100 & 15.62 & 3.95 & 1.71 & 0.0003 & 0.017 & 0.011  \\ \hline
R-PointHop*  & 3.58 & 1.89 & 0.58 & 0.0002 & 0.0150 & 0.0008  & 3.75 & 1.94 & 0.58 & 0.0002 & 0.0151 & 0.0008 \\ \hline
R-PointHop  & \bf{2.75} & \bf{1.66} & \bf{0.35} & \bf{0.0002} & \bf{0.0149} & \bf{0.0008}  & \bf{2.53} & \bf{1.59} & \bf{0.37} & \bf{0.0002} & \bf{0.0148} & \bf{0.0008} \\ \hline
\end{tabular}
\end{table*}

\begin{figure*}[htbp]
\centerline{\includegraphics[width=6.8in]{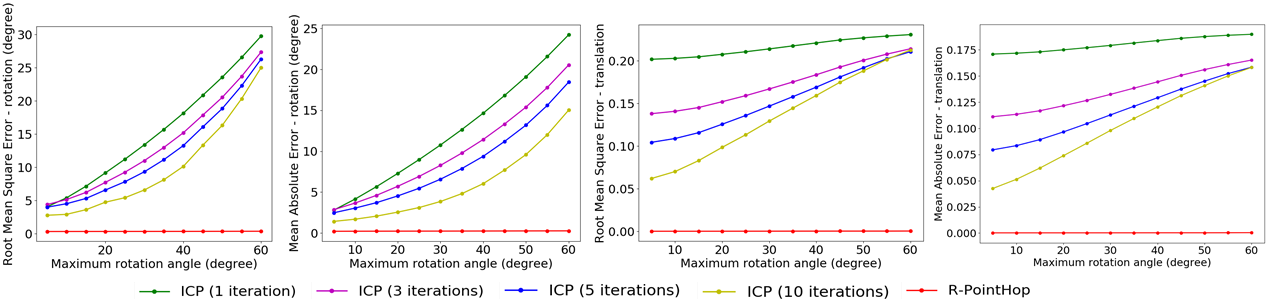}}
\caption{(From left to right) The plots of the maximum rotation angle
versus the root mean square rotation error, the mean absolute rotation
error, the root mean square translation error, and the mean absolute
translation error.}\label{fig:graph}
\end{figure*}

\begin{table}[htbp]
\centering
\caption{Registration on the Stanford Bunny dataset} \label{tab:Bunny}
\renewcommand\arraystretch{1.3}
\newcommand{\tabincell}[2]{\begin{tabular}{@{}#1@{}}#2\end{tabular}}
\resizebox{\columnwidth}{!}{
\begin{tabular}{c c c c c c c} \hline 
\bf Method & \tabincell{c}{\bf MSE\\(R)}  &  \tabincell{c}{\bf RMSE\\(R)}  
& \tabincell{c}{\bf MAE\\(R)} & \tabincell{c}{\bf MSE\\(t)}  &  \tabincell{c}{\bf RMSE\\(t)} 
& \tabincell{c}{\bf MAE\\(t)} \\ \hline 
\tabincell{c} 
ICP \cite{besl1992method}  & 177.35 & 13.32  & 10.72 & 0.0024 & 0.0492 & 0.0242 \\ \hline
Go-ICP \cite{yang2015go}  & 166.85  & 12.92   & 4.52 & 0.0018 & 0.0429 & 0.0282 \\ \hline
FGR \cite{zhou2016fast}  & 3.98  & 1.99  & 1.49  & 0.0397 & 0.1993 & 0.1658 \\ \hline
DCP \cite{wang2019deep}   & 41.45 & 6.44   & 4.78  & 0.0016 & 0.0406  & 0.0374  \\ \hline
R-PointHop  & \bf{2.21} & \bf{1.49} & \bf{1.09} & \bf{0.0013} & \bf{0.0361} & \bf{0.0269}  \\ \hline
\end{tabular}}
\end{table}

\subsubsection{Registration on Partial Data}\label{subsec:5.4}

Registration of partial point clouds is {common} in practical scenarios. We
considered the cases where the source and target have only a subset of
points in common. To generate a partial point cloud, we selected a point
at random and found its $N$ nearest neighbors. We set $N$ to be $3/4^{th}$
of the total number of points in the point cloud.
In our experiment, the initial point cloud has 1,024 points, and so the 
number of points in the partial point cloud is 768. The number overlapping 
points are between the source and target is thereby random between 512 and 768. 
The results of partial-to-partial 
registration are {presented} in Table \ref{tab:Result4}.
They are shown under two scenarios: 1) registration on unseen point
clouds and 2) registration on unseen classes.  R-PointHop gives the best
performance in the registration of partial data too. A critical element
in registering partial point clouds is to find correspondences between
overlapping points. R-PointHop handles it in the same way as those
presented in Secs.  \ref{subsec:5.1}-\ref{subsec:5.3} because of the use
of effective R-PointHop features to select good correspondences.
Furthermore, we show the effectiveness of using the ratio test to filter
out bad correspondences. The row of R-PointHop* in Table
\ref{tab:Result4} shows the errors when the ratio test is removed. The
errors are higher than those with the ratio test. Some results on
partial data registration are shown in Fig. \ref{fig:examples}, where
columns 3, 4 and 5 show the results where only the source is partial and
columns 6 and 7 show the results where both the source and the target
are partial. 

\begin{figure}[htbp]
\centerline{\includegraphics[width=3.4in]{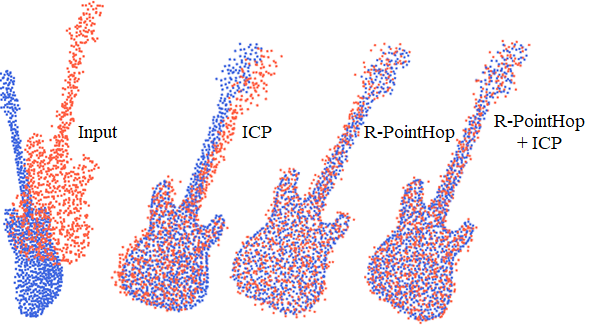}}
\caption{(From left to right) The source and target point clouds to be
aligned, registration with ICP only, with R-PointHop only, with
R-PointHop followed by ICP.}\label{fig:+ICP}
\end{figure}

\begin{figure*}[htbp]
\centerline{\includegraphics[width=7in,height=4.5in]{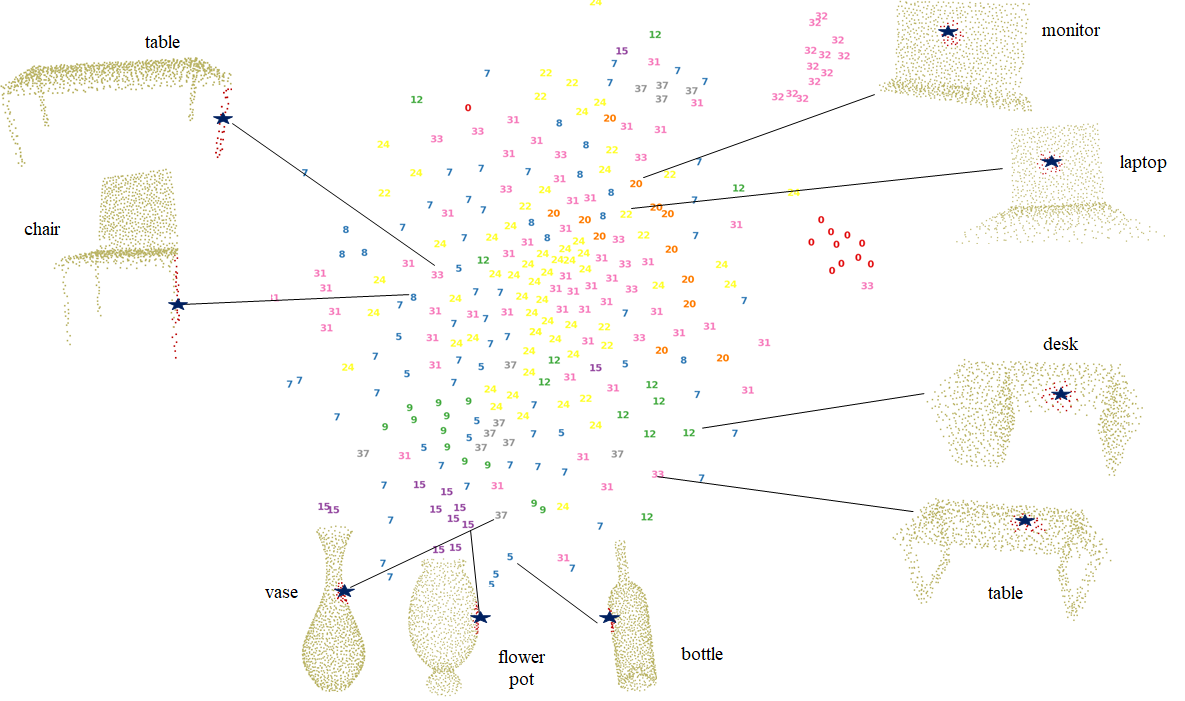}}
\caption{The t-SNE plot of point features, where a different number indicates
a different object class of points. Some points are highlighted and
their 3D location in the point cloud is shown. Features of points with a
similar local neighborhood are clustered together despite of differences
in their 3D coordinates.}\label{fig:tsne}
\end{figure*}

\subsubsection{Test on Real World Data}\label{subsec:5.5}

We {next tested} R-PointHop on 3D point clouds from the Stanford Bunny dataset
\cite{turk1994zippered}. It consists of 10 point cloud scans.
Typically, each scan contains more than 100k points.  In contrast with
the synthetic ModelNet40 dataset, it is a real world dataset.  We apply
a random spatial transformation to generate the source point clouds. For
registration, we select 2,048 points randomly so that they are evenly
spanned across the object. The R-PointHop derived from all 40 classes of
ModelNet40 is used for feature extraction.  For DCP, we use their model
trained on ModelNet40 and test on the Bunny dataset. We compare
R-PointHop with other methods and show the results in Table
\ref{tab:Bunny}.  One representative registration result is also shown
in Fig.  \ref{fig:Bunny}.  Table \ref{tab:Bunny} shows that R-PointHop
derived from ModelNet40 can be generalized to the Bunny dataset well.
In contrast, DCP does not perform so well on the Bunny dataset as
compared with ModelNet40.  We further experiment on point clouds from
the Stanford 3D scanning repository, which has a collection of several
categories of objects including Bunny, Buddha
\cite{curless1996volumetric}, Dragon \cite{curless1996volumetric}, {and}
Armadillo \cite{krishnamurthy1996fitting}. Some input scans and
their corresponding registered results are shown in Fig.
\ref{fig:Stanford3D}. 

\subsubsection{Local vs. Global Registration}\label{subsec:5.6}

ICP is local in nature and works only when the optimal alignment is
close to the initial alignment. In this case, R-PointHop can be used as
an initialization for ICP. That is, R-PointHop can {be used to obtain} the
initial global alignment, {after which} ICP can be used to {achieve} a tighter alignment. To
demonstrate this property, we plot the mean absolute error (MAE) and the
root mean squared error (RMSE) for rotation and translation against the
maximum rotation angle in Fig. \ref{fig:graph}. As shown in the figure,
as the maximum rotation angle increases, the MAE and the RMSE for ICP
increase steadily. In contrast, the RMSE and the MAE of R-PointHop are
very stable, reflecting the global registration power of R-PointHop. In
Fig.  \ref{fig:+ICP}, we show three registration results: 1) using ICP
alone, 2) using R-PointHop alone, and 3) R-PointHop followed by ICP. We
can obtain slightly better results in the third case as compared to the second case. However, without {initializing with
R-PointHop,} ICP fails to align well. 

\subsubsection{3D Descriptor}\label{subsec:5.7}
\begin{figure}[htbp]
\centerline{\includegraphics[width=3.4in]{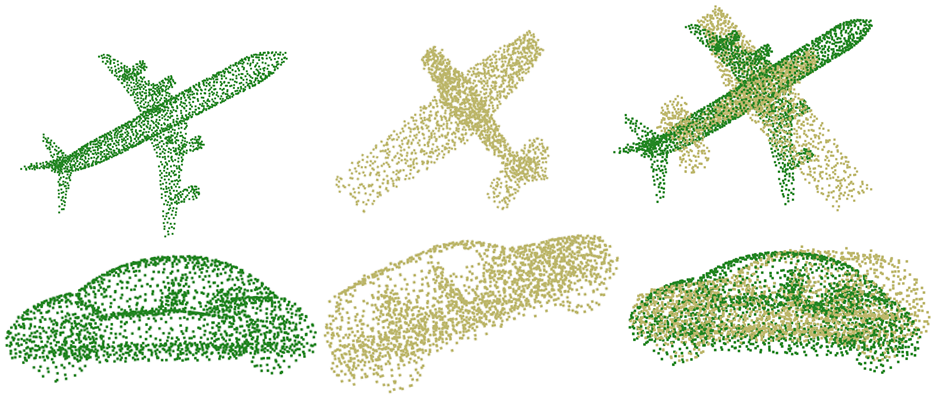}}
\caption{Registration of two point cloud models, where the first two
columns are input point clouds and the third column is the output
after registration.}\label{fig:different}
\end{figure}

{Fig. \ref{fig:tsne} shows} the t-SNE plot of some point features obtained by R-PointHop. It is observed that the points of a similar local
structure are close to each other, irrespective of their spatial
locations in the 3D point cloud model as well as whether they belong to
the same object or the same class. To give an example, we show two point
cloud models of a table and a chair in the left. The points on their
legs have a similar neighborhood structure and their features are closer
in the t-SNE embedding space. This demonstrates the capability of
R-PointHop as a general 3D descriptor. As an application, we show the
registration of two different objects of the same object class in Fig.
\ref{fig:different}, which has two different airplanes and cars.
Although the objects are different, we can still align them reasonably
well. This is because points in similar semantic regions are selected
as correspondences. Apart from 3D correspondence and registration, the
3D descriptor can be used for a variety of applications such as point
cloud retrieval, which can be a future extension of this work. 

\subsection{Ablation Study}\label{subsec:5.8}

The effects of {the} model parameters, ratio test and {use of} RANSAC on {the} object registration performance are {discussed} in this
subsection.  We report the mean absolute errors using different input
point numbers and neighborhood sizes of LRF in the first section of
Table \ref{tab:Ablation-ModelNet40}.  The error values are comparable
for different numbers {of input points} and LRF neighborhood sizes. The
performance slightly drops when 512 points are used. Hence, we fix 2048
points and set the neighborhood size of LRF to 128 in following
experiments.  Next, we consider various degree of partial overlaps and
the effect of adding noise of three levels in the second and the third
sections of Table \ref{tab:Ablation-ModelNet40}, respectively.  For
partial registration, the error increases as the maximum overlapping
region decreases. We see consistent improvement with the ratio test and
RANSAC. Similarly, the performance improves after inclusion of the ratio
test and RANSAC for registration with noise.

\begin{table*}[htbp]
\centering
\caption{Ablation study on object registration.} \label{tab:Ablation-ModelNet40}
\renewcommand\arraystretch{1.3}
\newcommand{\tabincell}[2]{\begin{tabular}{@{}#1@{}}#2\end{tabular}}
\begin{tabular}{c c c c c c c c} \hline 
{\bf Input points} & \tabincell{c}{{\bf LRF}}  &  \tabincell{c}{{\bf Partial overlap}}  & \tabincell{c}{{\bf Noise std.}} & \tabincell{c}{{\bf Ratio test}} & \tabincell{c}{{\bf RANSAC}} &
\tabincell{c}{{\bf MAE(R)}} & \tabincell{c}{{\bf MAE(t)}} \\ \hline 
{1024}  & {64} &   &  & & & {0.24} & {0.000301} \\ 
{1024}  & {32} &  &  & & & {0.25} & {0.000314} \\ 
{2048}  & {128} &  &  & {\checkmark} & & {0.24} & {0.000297} \\ 
{2048}  & {64}  &  &  & {\checkmark} & & {0.24} & {0.000300}\\ 
{1024}  & {64} &  &  & {\checkmark} & & {0.24} & {0.000295} \\
{512}   & {32} &  &  & {\checkmark} & & {0.29} & {0.000546} \\  \hline

{1536}  & {96} & {75\%}  &  & & & {0.56} & {0.000856} \\ 
{1024}  & {64} & {50\%} &  & & & {2.41} & {0.001340} \\ 
{512}  & {32} & {25\%} &  & & & {8.67} & {0.031237} \\ 
{1536}  & {96}  & {75\%}  &  & {\checkmark} & {\checkmark} & {0.31} & {0.000824} \\ 
{1024}   & {64} & {50\%} &  & {\checkmark} & {\checkmark} & {0.87} & {0.001339} \\ 
{512}  & {32} & {25\%} &  & {\checkmark} & {\checkmark} & {6.69} & {0.031202} \\ \hline

{2048}  & {128} &   & {0.01} & & & {0.99} & {0.003752} \\ 
{2048}  & {128} &  & {0.05} & & & {1.43} & {0.004138} \\ 
{2048}  & {128} &  & {0.1} & & & {2.81} & {0.007123} \\ 
{2048}  & {128}  &   & {0.01} & {\checkmark} & {\checkmark} & {0.88} & {0.003711} \\ 
{2048}   & {128} &  & {0.05} & {\checkmark} & {\checkmark} & {1.37} & {0.004122} \\ 
{2048}  & {128} &  & {0.1} & {\checkmark} & {\checkmark} & {2.74} & {0.007093} \\ \hline
\end{tabular}
\end{table*}

To gain further insights into the difference {between}
simple object point clouds and complex indoor point clouds, we
remove the surface normal and geometric features and use only point
coordinates for indoor registration. This leads to a sharp decrease in
performance, {to an} average recall and precision {of} 0.39 and 0.19,
respectively. Clearly, the use of point coordinates is not sufficient
for {the registration of} complex indoor point clouds. We also reduce the LRF neighborhood
size for indoor point clouds and see whether 64 or 128 neighbors could
give similar performance as observed in object registration. Again,
there is some performance degradation, and the best results are achieved
with 256 neighbors. This is attributed to the fact that the more the
number of points used to find the LRF, the more stable the local PCA
against small perturbations and noise. In other words, the optimal point
attributes and the hyper-parameter settings are different for
{registering} object and indoor {scene} point clouds.

\subsection{Toward Green Learning}\label{subsec:5.9}

One shortcoming of deep learning methods is that they tend to have a
large model size, which make them difficult to deploy on mobile devices.
{Moreover}, recent studies indicate that training deep learning models has a large carbon footprint. Along with the environmental impact,
expensive GPU resources are needed to successfully train these networks
in reasonable time. The need to search {for} an environmental friendly green
solution to different AI tasks, or green AI \cite{schwartz2019green}, is
on the rise.  Although the use of efficiency (training time, model size
etc.) as an evaluation criterion along with the usual performance
measures was emphasized in \cite{schwartz2019green}, no specific green
models were presented. 

R-PointHop offers a green solution in terms of a smaller model size 
and training time as compared with deep-learning-based methods. We
trained PointNetLK and DCP methods using the open source codes provided
with the default parameters set by authors. We compare the training
complexity below.
\begin{itemize}
\item DCP took about 27.7 hours to train using eight NVIDIA Quadro
M6000 GPUs. 
\item PointNetLK took approximately 40 minutes to train one epoch
using one GPU while the default training setting is 200 epochs. Thus,
the total training time was 133.33 hours.
\item R-PointHop took only 40 minutes to train all model parameters
using {an} Intel(R) Xeon(R) CPU E5-2620 v3 at 2.40GHz.
\end{itemize}

The inference time of all methods {was} comparable.  However, since ICP,
Go-ICP, SPA, and PointNetLK are iterative methods, their inference time
is a function of the iteration number. We observe that the required
iteration number varies from model to model.  

The model size of R-PointHop is only 200kB compared {to} 630kB {for}
PointNetLK and 21.3MB of DCP. The use of transformer makes the model
size of DCP significantly larger.  Although the model free methods are
most favorable in terms of model sizes and training time, their
registration performance is much worse. Thus, R-PointHop offers a good
balance when all factors are considered. 

\section{Discussion}\label{sec:discussion}

\subsection{Role of Supervision}

{To determine} whether the performance gain using
supervised deep learning is due to large unlabeled data, data
labeling, or both, we split experiments on
ModelNet40 into two parts (i.e., tests on seen and unseen object
classes) as a case study. Some supervised learning methods {performed}
poorer on unseen classes (see Tables \ref{tab:Result1}, \ref{tab:Result2}, and \ref{tab:Result3}), which indicates that they learn object
categories indirectly, {even though their} supervision {uses} ground truth
rotation/translation values without class labels. This behavior is not
surprising since the two benchmark methods, PointNetLK and Deep Closest
Point (DCP). are derived from PointNet and DGCNN, respectively,
which were designed for point cloud classification. In contrast, our
feature extraction is rooted in PointHop, which is unsupervised and
task-agnostic. Our model does not know the downstream task. Hence, it
can generalize well to unseen classes. To show this point furthermore,
we use the R-PointHop model learned from ModelNet40 and evaluate it on
the Stanford bunny model. Its performance gap is smaller than those of
supervised learning methods.  These experiments indicate that the
performance gain of supervised learning methods is somehow limited to
similar instances of point clouds that the models have already seen and
their generalization capability to unseen classes is weaker.

\subsection{Limitations of R-PointHop}

In general, we see that R-PointHop works extremely
well for the object registration case and also matches the performance
with PPFNet for indoor registration. However, there exist some recent
exemplary networks (e.g., \cite{gojcic2019perfect}) that have a higher
recall on the 3DMatch dataset. The eight octant partitioning operation
in the feature construction step fails to encode better local structure
information for point clouds from this dataset. Since R-PointHop is
based on successive aggregation of local neighborhood information, an
initial set of attributes that captures better local neighborhood
structure can help improve the performance. One such choice could be 
the FPFH descriptor.

For ModelNet40, we see that the performance {of R-PointHop} degrades
when the amount of overlap reduces {or the amount of} noise increases. One
reason is the stability of LRF. It is observed that a larger
neighborhood number tends to compensate for noise and surface variations
in our experiments on the 3DMatch dataset. However, when the number of
points in a dataset is small, we cannot opt for more points in finding
LRF. Although RANSAC offers a more robust solution, a
fine alignment step {may be necessary}.  That is, some ICP iterations {may achieve} a tighter alignment.

\section{Conclusion and Future Work}\label{sec:conclusion}

An unsupervised 3D registration method, called R-PointHop, was proposed
in this work. R-PointHop extracts point features of varying neighborhood
sizes in a one-pass manner, where the neighborhood size grows as the {number of} hop increases.  Features extracted by R-PointHop are invariant with
respect to rotation and translation due to the use of the local
reference frame (LRF). {This enables} R-PointHop {to find}
corresponding pairs accurately in presence of partial point clouds and
larger rotation angles. It was shown by experimental results that
R-PointHop offers the state-of-the-art performance in point cloud
registration. Furthermore, its training time and model size are less
than those of deep learning methods by an order of magnitude. 

It is worth {noting} that R-PointHop does not follow the
end-to-end optimization framework as adopted by deep learning methods
nowadays. This choice makes R-PointHop a green solution. Also, it is
typical that supervised learning methods outperforms unsupervised
learning methods. {But}, our work shows that ground truth transformations are
not necessary in the point cloud registration problem. 

It appears that the usage of extracted features is not confined to the
registration problem.  These features may be used as a general 3D point
descriptor.  We would like to explore this idea and check the usefulness
of R-PointHop as a 3D descriptor on large scale point clouds in the
future. It is also interesting to extend the proposed solution to the
more challenging task of LiDAR odometry. The incremental motion of the
object can potentially be estimated using R-PointHop by finding the
point correspondences between consecutive point cloud scans.
Furthermore, another application of the proposed
solution can be simultaneous object retrieval and registration, where a
similar object to a query object can be retrieved from a database and
aligned to it.

\section*{Appendix \\ Channel-Wise Saab transform}

The Saab transform \cite{kuo2019interpretable} is
derived from the Principal Components Analysis (PCA). It adds a bias
term {that} annihilates the need {for} a non-linear activation function when
multiple PCA transforms are cascaded. The transform can be further
modified by applying the Saab transform to each channel separately by
exploiting the channel decoupling property. It is termed as the
channel-wise (c/w) Saab transform \cite{zhang2020pointhop++}. The model
size of the c/w Saab transform is significantly smaller than that of the
Saab transform. The c/w Saab transform is described below.

Initially, the standard Saab transform is performed
in the first hop. The output Saab coefficients are given by
\begin{equation}
    y_k = \sum_{n=0}^{N-1} a_{k,n}v_n + b_k = \mathrm{a_k^Tv}+b_k, 
\ k = 0,1,\cdots,K-1,
\end{equation}
where, $\mathrm{v}=[v_0,v_1,\cdots,v_{N-1}]^T$ is
the $N$-dimensional input vector, $y_k$ is the $k$-th Saab coefficient,
$\mathrm{a_k}=[a_{k,0},a_{k,1},\cdots,a_{k,N-1}]^T$ is the weight
vector, and $b_k$ is the bias term.

For the DC Saab filter $(k=0)$, the weight is given by
\begin{equation}
\mathrm{a_0} = \frac{1}{\sqrt{N}}[1,1,\cdots,1]^T.
\end{equation}
The DC component is obtained by projecting $v$ onto 
the DC filter as
\begin{equation}
\mathrm{v_{DC}}=\frac{1}{\sqrt{N}}\sum_{n=0}^Nv_n.
\end{equation}
The AC component is given by
\begin{equation}
\mathrm{v_{AC}}=\mathrm{v}-\mathrm{v_{DC}}.
\end{equation}
The AC Saab filters a$_1, \cdots , \mathrm{a}_K$ are
{obtained} by performing PCA on the AC component v$_{AC}$. The first
$K-1$ principal components are selected as the AC filters. Finally, the
bias term $b_k$ is selected such that
\begin{equation}
b_k \geq \underset{\mathrm{v}}{\max}\|\mathrm{v}\|, \quad k=0, \cdots , K-1.
\end{equation}
This choice of $b_k$ guarantees that $y_k$ is always
non-negative, thereby removing the need of a non-linear activation like
ReLU.

Since the Saab transform is a variant of PCA, the
Saab coefficients are weakly correlated. Due to the weak spectral
correlations, the joint spatial-spectral tensor of dimension $K$ at the
input of the second hop {is} decomposed into $K$ spectral tensors. Later, {the}
Saab transform is performed on each of the $K$ spectral channels
separately. Due to this nature, it is called the channel-wise (c/w) Saab
transform.

The multi-hop feature learning process then leads to
the feature tree representation, where each node of the tree corresponds
to one spectral component. The spectral components at the output of the
first hop are the children of the root node. Every node in the tree is
associated with an energy. The energy of every child node is the product
of the energy of its parent node and its normalized energy with respect
to all its siblings. An energy threshold $T$ is a hyperparameter that
decides whether the node goes to the next hop or not. Nodes with {energies}
greater than $T$ are passed on to the next hop. These nodes are called
as intermediate nodes. {Meanwhile,} the nodes with {energies} less than $T$ are collected
as leaf nodes. Each leaf node represents a single feature dimension and
the components of all the leaf nodes are concatenated to {obtain} the output
feature.

\section*{Acknowledgment}

This work was supported by a research grant from Tencent Media Lab. 

\ifCLASSOPTIONcaptionsoff
  \newpage
\fi

\bibliographystyle{IEEEtran}
\bibliography{ref}

\begin{IEEEbiography}[{\includegraphics[width=1in,height=1.25in, 
clip,keepaspectratio]{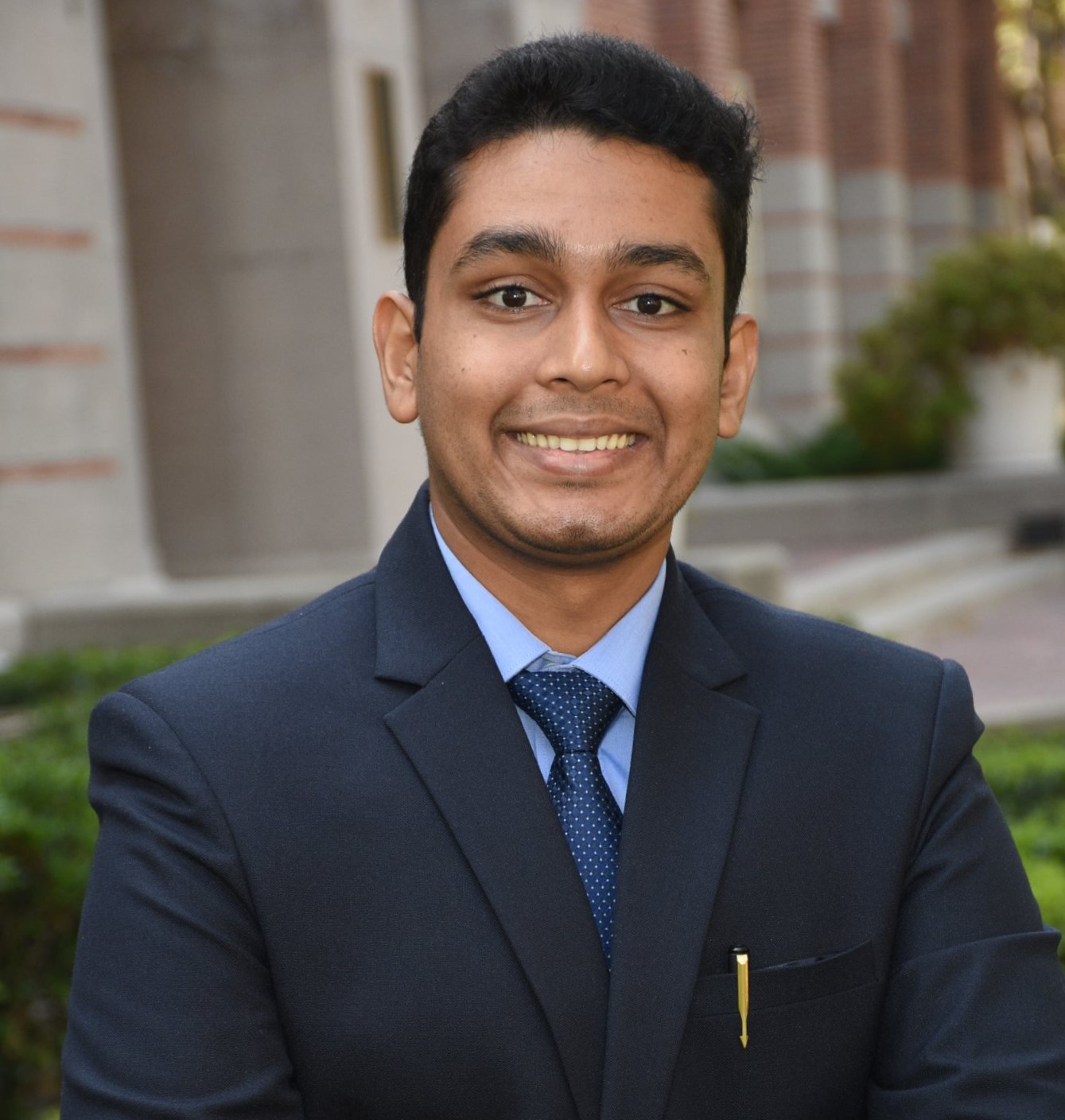}}]{Pranav Kadam} received his MS degree in Electrical Engineering from the University of Southern California, Los Angeles, USA in 2020, and the Bachelor’s degree in Electronics and Telecommunication Engineering from Savitribai Phule Pune University, Pune, India in 2018. He is currently pursuing the PhD degree in Electrical Engineering from the University of Southern California. He is actively involved in research and development of methods for point cloud analysis and processing. His research interests include 3D computer vision, machine learning, and perception.
\end{IEEEbiography}

\begin{IEEEbiography}[{\includegraphics[width=1in,height=1.25in, 
clip,keepaspectratio]{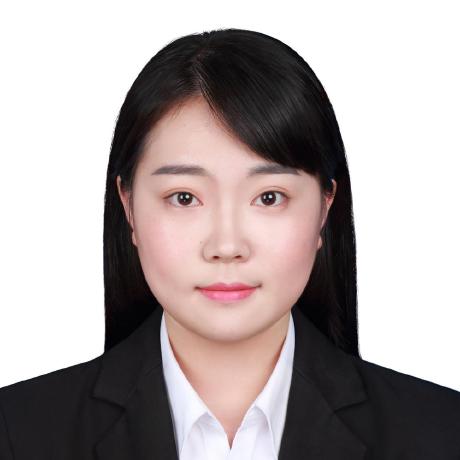}}]{Min Zhang}
received her B.E. degree from the School of Science, Nanjing University
of Science and Technology, Nanjing, China and her M.S. degree from the
Viterbi School of Engineering, University of Southern California, Los
Angeles, US, in 2017 and 2019, respectively. She is currently working
toward the Ph.D. degree from University of Southern California. Her
research interests include 3D Vision and Machine Learning. 
\end{IEEEbiography}

\begin{IEEEbiography}[{\includegraphics[width=1in,height=1.25in,
clip,keepaspectratio]{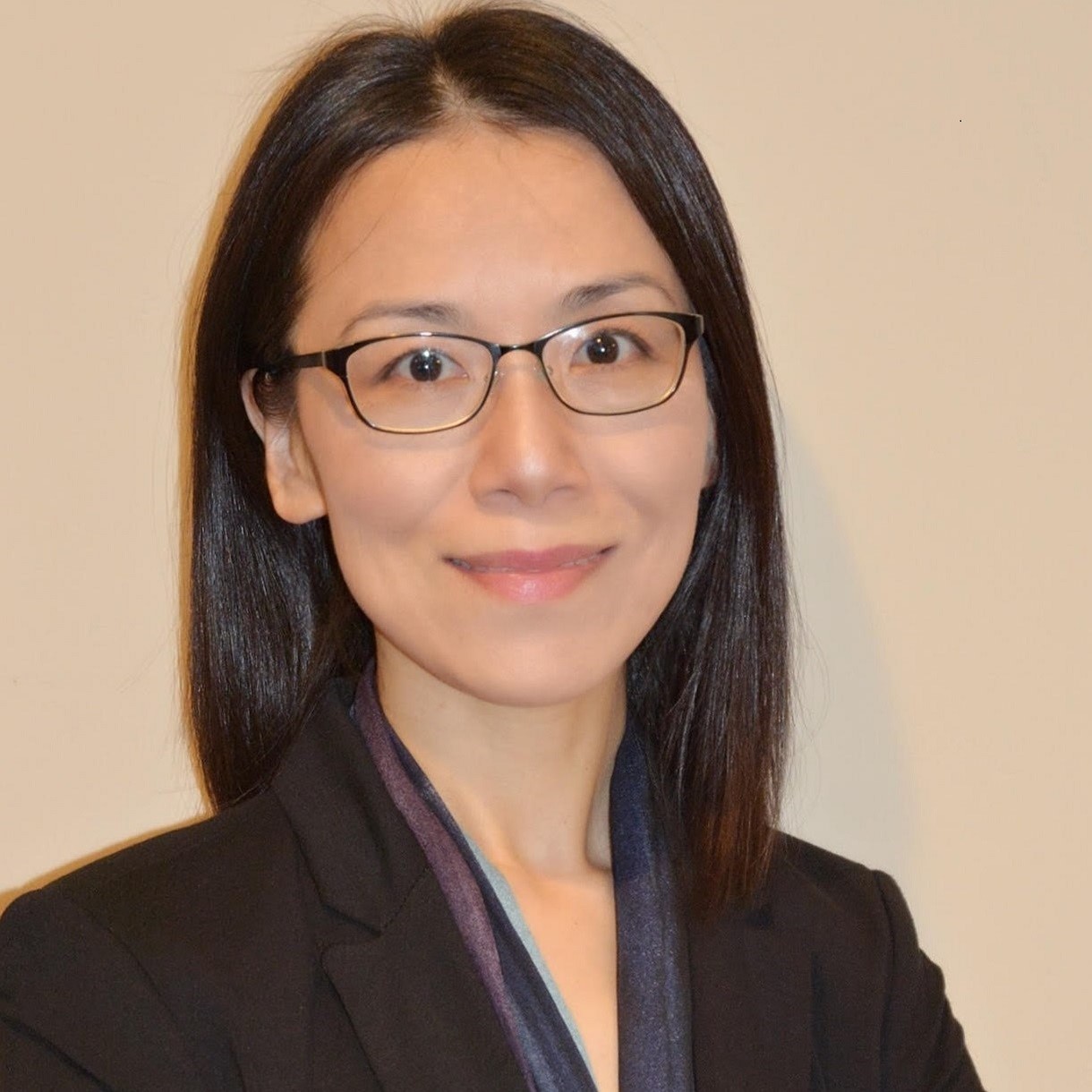}}]{Shan Liu} (M’01-SM’11) received the B.Eng. degree in electronic engineering from Tsinghua University, the M.S. and Ph.D. degrees in electrical engineering from the University of Southern California, respectively. She is a Distinguished Scientist at Tencent and General Manager of Tencent Media Lab. She was formerly Director of Media Technology Division at MediaTek USA. She was also formerly with MERL and Sony, etc. She has been an active contributor to international standards for more than a decade and has numerous technical proposals adopted into various standards, such as VVC, HEVC, OMAF, DASH, MMT and PCC. She was an Editor of H.265/HEVC SCC and H.266/VVC standards. She is an APSIPA Distinguished Industry Leader and a vice chair of IEEE Data Compression Standards Committee. She received the Best AE Award from IEEE TCSVT in 2019 and 2020. She holds more than 300 granted US patents. Her research interests include audio-visual, volumetric, immersive and emerging media compression, intelligence, transport and systems. 
\end{IEEEbiography}

\begin{IEEEbiography}[{\includegraphics[width=1in,height=1.25in,
clip,keepaspectratio]{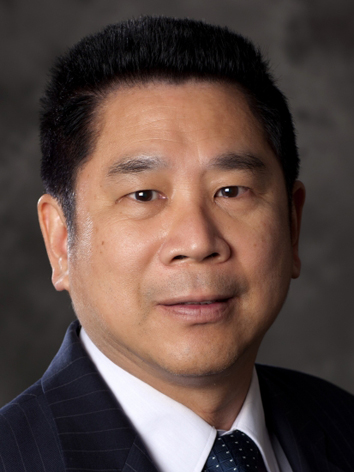}}]{C.-C. Jay
Kuo} (F’99) received the B.S. degree in electrical engineering from
the National Taiwan University, Taipei, Taiwan, in 1980, and the M.S.
and Ph.D. degrees in electrical engineering from the Massachusetts
Institute of Technology, Cambridge, in 1985 and 1987, respectively. He
is currently the Director of the Multimedia Communications Laboratory
and a Distinguished Professor of electrical engineering and computer
science at the University of Southern California, Los Angeles. His
research interests include digital image/video analysis and modeling,
multimedia data compression, communication and networking, and
biological signal/image processing. He is the coauthor of about 280
journal papers, 940 conference papers and 14 books. Dr. Kuo is a Fellow
of the American Association for the Advancement of Science (AAAS) and
The International Society for Optical Engineers (SPIE). 
\end{IEEEbiography}

\end{document}